\documentclass{article}

\usepackage[final]{neurips_2024}

\usepackage[utf8]{inputenc} 
\usepackage[T1]{fontenc}    
\usepackage{hyperref}       
\usepackage{url}            
\usepackage{booktabs}       
\usepackage{amsfonts}       
\usepackage{nicefrac}       
\usepackage{microtype}      
\usepackage{xcolor}         
\usepackage{amsmath}
\usepackage{graphicx}
\usepackage{subcaption}

\title{Learning Modular Exponentiation with Transformers}
\author{%
  David Demitri Africa\thanks{Equal contribution. Correspondence: \texttt{\{dda28,smk78,tss52\}@cam.ac.uk}}  ~~~   Sara M. Kapoor\footnotemark[1] ~~~  Theo Simon Sorg\footnotemark[1] \\
  ~~~ \textbf{Challenger Mishra}\\
  Department of Computer Science and Technology \\
  University of Cambridge \\
  }

\begin{document}

\maketitle

\begin{abstract}
Modular exponentiation ($a^b \equiv d \bmod c$) is crucial to number theory and cryptography, yet remains largely unexplored from a mechanistic interpretability standpoint. We train compact 4‑layer encoder–decoder Transformers to predict $d$ and analyze how they come to solve the task. We compare principled sampling schemes for $(a,b,c,d)$, probe the learned token embeddings, and use causal interventions (activation patching) to localize the computation inside the network. Sampling $a$ and $b$ log‑uniformly (reciprocal sampling) removes severe output imbalance and yields large accuracy gains, with abrupt, synchronized jumps in accuracy that simultaneously cover families of related moduli (e.g., multiples of 23). Causal analysis shows that, on instances without reduction ($c > a^b$), a small circuit consisting only of final‑layer attention heads reproduces full‑model behavior, indicating functional specialization. These results suggest that Transformers can internalize modular arithmetic via compact, specialized circuits, and that data distribution strongly shapes both learning dynamics and generalization.
\end{abstract}

\section{Introduction}
Modular exponentiation is fundamental in cryptographic algorithms such as RSA \citep{rivest1978method} and Diffie-Hellman key exchange \citep{1055638}. Despite its mathematical simplicity, modular exponentiation presents significant challenges for machine learning models, given the non-linear and cyclic nature of the operation. Prior work by \citet{charton2024learning} explored how transformers can learn arithmetic functions such as the greatest common divisor (GCD), showing that transformers implicitly uncover algorithmic structures in arithmetic tasks. Computationally, modular exponentiation is efficiently solvable, as the decision problem “is $a^b \equiv d \ (\bmod\ c)$?” lies in P since repeated squaring computes $a^b \bmod c$ with $O(\log b)$ modular multiplications (overall bit‑complexity $O(M(n)\log b)$ for $n$‑bit inputs) \citep{knuth1969vol, von2003modern, menezes2018handbook}.

Given the interaction between exponentiation and modulo reduction, modular exponentiation may present nontrivial number-theoretic patterns not present in simpler tasks. Transformer-based models have not been systematically investigated for modular exponentiation, particularly from a mechanistic interpretability perspective. As such, uncovering novel patterns in modular exponentiation may also have knock-on effects in the study of diophantine geometry and modularity. In general, machine driven mathematical discovery has been a rich area of exploration, with large datasets available for machine learning exploration in conjecture generation \citep{davies2021advancing, he2024murmurations, wang2025discoveryunstablesingularities}.

In this work, we train transformer models to perform modular exponentiation, design novel sampling methods to capture the statistical characteristics of modular arithmetic, and analyze how these sampling strategies influence model learning dynamics and generalization. We then use mechanistic interpretability methods to understand how models represent number-theoretic information internally. We also observe evidence of grokking as the model learns individual multiples.

\section{Related Work}
\label{sec:related_work}

Mechanistic interpretability seeks to uncover computational strategies internalized by neural networks at the level of individual components such as attention heads and weight matrices. \citet{olah2020zoom} introduced the concept of \textit{circuits}, coherent subgraphs corresponding to meaningful algorithmic operations, with subsequent work by \citet{elhage2021mathematical} and \citet{wang2022interpretability} expanding this framework to dissect transformer internal structures. Recent mechanistic interpretability has been applied specifically to arithmetic reasoning, with \citet{quirke2024understanding} discovering dedicated attention heads mirroring human arithmetic algorithms and \citet{stolfo2023mechanistic} using causal mediation analysis to reveal interpretable arithmetic pathways. Machine learning approaches to modular arithmetic have focused on simpler operations like modular addition \citep{gromov2023grokking, saxena2024modular}, with \citet{gromov2023grokking} demonstrating neural networks learning modular addition through \textit{grokking}, sudden generalization from memorization to algorithmic understanding initially documented by \citet{power2022grokking}. \citet{nanda2023progress} introduced internal progress measures to identify circuits responsible for emergent arithmetic capability, while \citet{doshi2024modular} studied grokking in modular polynomial arithmetic. Our work extends these explainability techniques to the previously unexplored domain of modular exponentiation.

\section{Sample Generation and Model Training}
\label{sec:samplegen}
Following \citet{charton2024learning}, we experiment with various sampling methods for modular exponentiation. Modular exponentiation requires sampling integers $a,b,c \in \mathbb Z$ and outcome $d \in \mathbb Z$ such that
\begin{equation}\label{eq:dmod}
    a ^ b \equiv d \mod c
\end{equation}
We sample $a,b,c$ and sometimes $d$ such that tuples $(a,b,c,d)$ follow a certain underlying distribution, with the maximum integer to sample set to $M = 10^6$. We find that different distributions lead to varying learning dynamics and absolute accuracies, with greater imbalance leading to lower accuracy.

\paragraph{Uniform operands} samples $c \in [1,100]$ and $a,b \in [0, M]$ uniformly, then computes $d$. This creates severe class imbalance with $d$ heavily skewed toward small values, preventing the model from learning larger outcomes.

\paragraph{Uniform outcomes} mitigates this by sampling $d$ uniformly, then rejection sampling $(a,b,c)$ such that \eqref{eq:dmod} holds with constraint $c > d$.

\paragraph{Reciprocal operands} samples $a,b$ log-uniformly by sampling $\ln(a), \ln(b) \sim \mathcal U(1, \ln(M+2))$, computing $a = \lfloor e^{\ln(a)} \rfloor, b = \lfloor e^{\ln(b)} \rfloor$, then shifting by 1 to include $a,b = 0$. This yields the discrete probability distribution:
\begin{equation}
    \mathcal P(n, 0, M) = \begin{cases}0 & n < 0 \text{ or } n > M\\
    \frac{\ln(n + 2) - \ln(n+1)}{\ln(M+2)} & 0 \le n \le M
    \end{cases}
\end{equation}

We detail the proof for this in Appendix \ref{append:operanddistribution}. In training, we test four combinations: uniform/reciprocal operands × computed/uniform outcomes. For comparability with \citet{charton2024learning} train four-layer encoder-decoder transformers with embedding dimension 256, eight attention heads, batch size 256, and learning rate $10^{-4}$ with the Adam optimizer \citep{kingma2014adam}. Each epoch uses 300,000 generated samples.

\paragraph{Integer representations} Since transformers operate on discrete tokens, and the range of integers up to M is too large to use as a vocabulary for the small transformers trained here, we follow \citet{charton2024learning} and represent integers using base $B$ digits in the template:
\begin{equation*}
    \mathrm{V3~} + a_1 \ldots a_n + b_1 \ldots b_n + c_1 \ldots c_n + d_1 \ldots d_n
\end{equation*}
For example, $750178^{996884} \equiv 1 \mod 95$ becomes $\mathrm{V3} +750~178~ +996~884~ +95~ +1$ in base 1000. This string is constructed using the samples generated during training.

\paragraph{Evaluation} We test all four sampling methods for 2500 epochs using base 1000, with validation set (uniform operands, computed outcomes) and test set (uniform operands, uniform outcomes) to assess performance across distributions. We also test four bases on the best-performing reciprocal operands setting to study the impact of base choice.


\section{Results}

\begin{figure}
    \centering
    \includegraphics[width=0.95\linewidth]{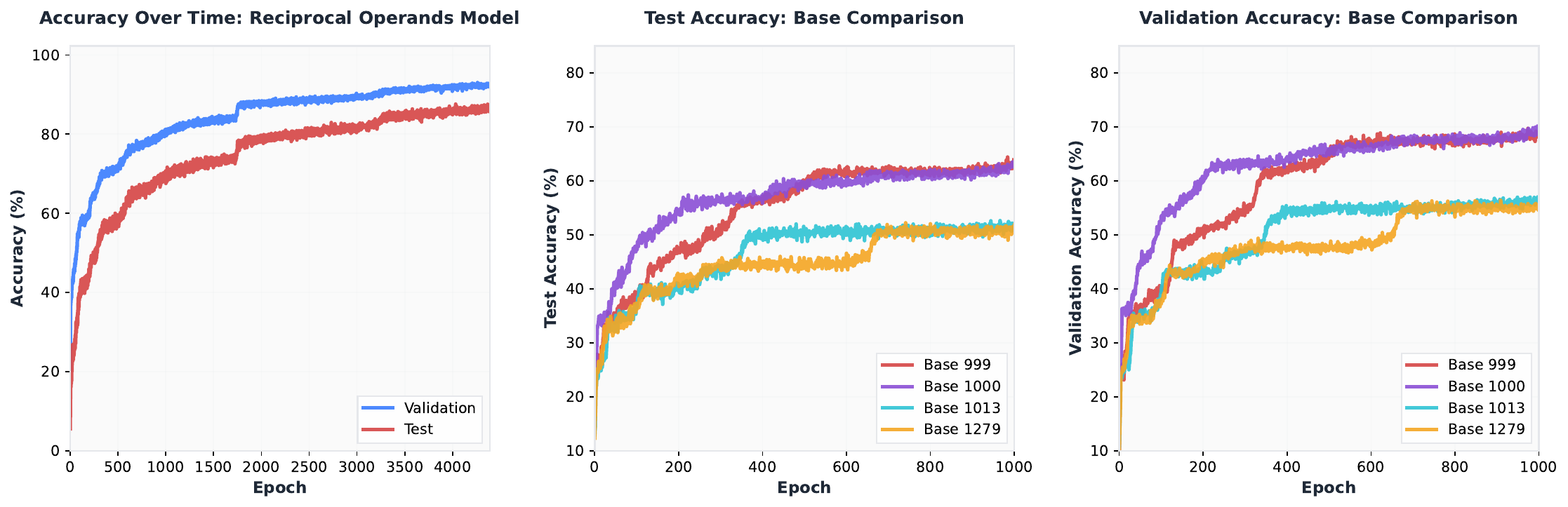}
\caption{\textit{Left}: Validation and test accuracy over 3000 epochs for the reciprocal operands model. Reciprocal sampling (log-uniform distribution of operands) enables effective learning of modular exponentiation, with validation accuracy reaching $\sim$84\% and test accuracy $\sim$80\%. \textit{Middle}: Test accuracy comparison across four numerical bases over 1000 epochs. Composite bases (999, 1000) substantially outperform prime bases (1013, 1279), with bases 999 and 1000 reaching $\sim$60\% accuracy compared to $\sim$49\% for prime bases. \textit{Right}: Validation accuracy shows the same pattern, confirming that the composite base advantage generalizes across both evaluation sets. Base choice significantly impacts learning dynamics and final performance.}
    \label{fig:benford}
    \label{fig:basecomparison}
\end{figure}

Transformer models successfully learn modular exponentiation, with the best performing model reaching over 80\% test accuracy after 3000 epochs (\autoref{fig:benford}).


\paragraph{Performance on modular exponentiation.} Reciprocal operands sampling yields dramatically better performance than uniform operands (\autoref{tab:samplingresults}), resolving class imbalance without requiring uniform outcomes. In the following, samples are generated using reciprocal operands if not explicitly mentioned otherwise. Prime number bases clearly perform worse, although it is not clear why\footnote{We suspect that composite bases like $1000 = 2^3 × 5^3$ may facilitate learning by aligning with divisibility structure in modular arithmetic, though further investigation is needed.}. However, when comparing the two prime numbers or composite numbers with each other, there is no clear advantage. Both prime numbers and the composite numbers perform equally well, respectively. The subsequent experiments were conducted with base 1000, as bases 1000 and 999 achieve similar accuracy.


\begin{table}[h]
    \centering
    \begin{tabular}{c|cc}
         & Computed & Uniform \\
         Uniform operands& 13.17& 28.14\\
         Reciprocal operands& \textbf{80.39} & 79.16
    \end{tabular}
    \caption{Test accuracy (\%) for sampling methods.}
    \label{tab:samplingresults}
\end{table}

\paragraph{Evaluating deterministic predictions.} Following \citet{charton2024learning}, we analyze deterministic mispredictions. The model consistently predicts 19 instead of the correct 91 across all 10,012 samples (512 distinct ones) with target 91. When we control for duplicates and scale up samples with fixed $d=91$, prediction 1 becomes most common for target 91, with 19 persisting for infrequent $a,b,c$. We hypothesize that the 1 might serve as a fall-back mechanism for our model, which would be relevant for unseen and rare data.

\begin{figure}
    \centering
    \includegraphics[width=0.9\linewidth]{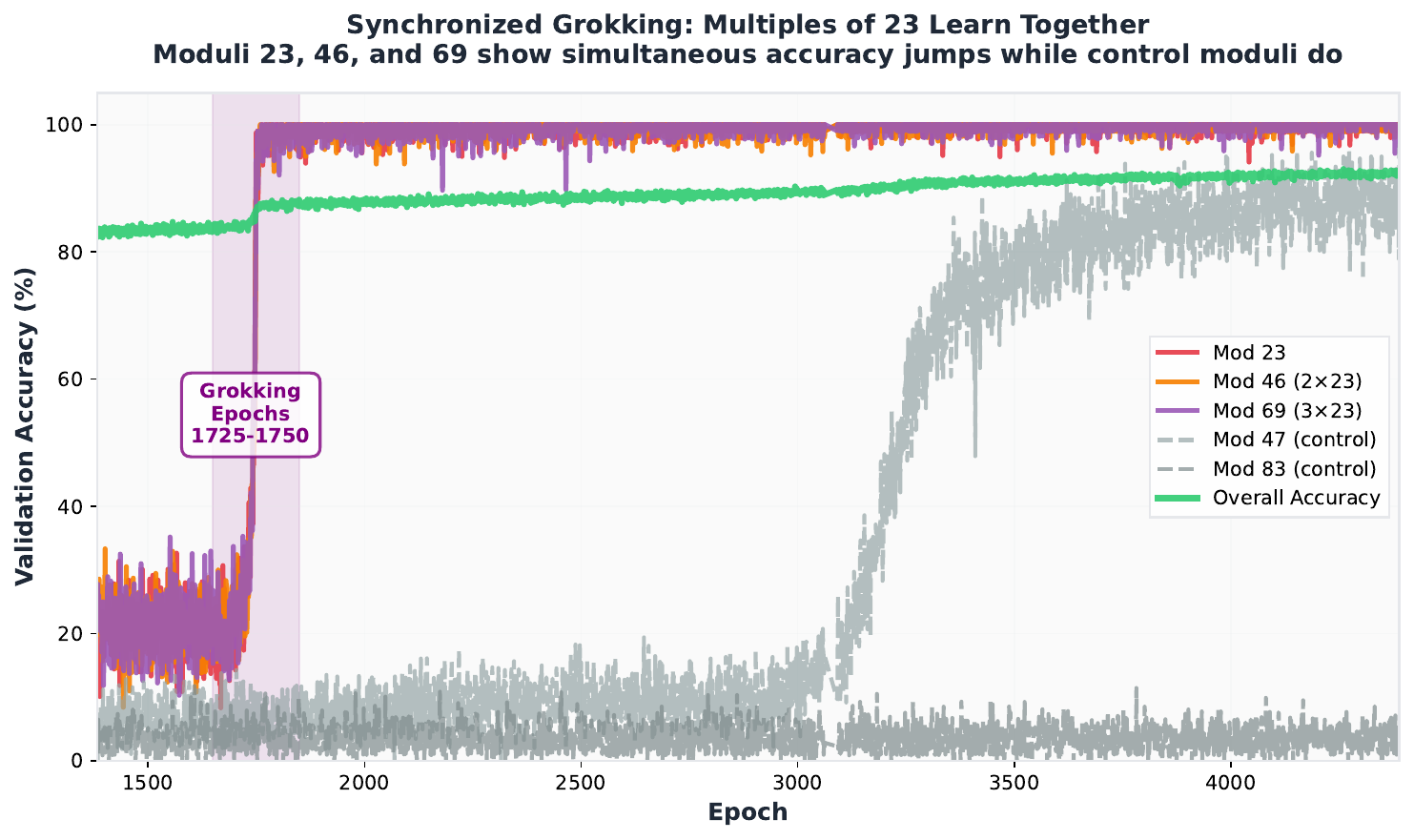}
\caption{Synchronized grokking for multiples of 23 during epochs 1725--1750 (highlighted region). Moduli 23, 46 ($2\times23$), and 69 ($3\times23$) exhibit simultaneous accuracy jumps from $\sim$20\% to near-perfect performance, demonstrating that the transformer discovers and exploits mathematical relationships between related moduli. Control moduli 47 and 83 (not multiples of 23) exhibit a different learning pattern with gradual improvement, while overall accuracy remains high ($\sim$83\%) throughout training.}
    \label{fig:validcomparison}
\end{figure}

\paragraph{Learning dynamics analysis.} We observe significant performance surges between epochs 1725-1750, with simultaneous accuracy increases from 20\% to 100\% for multiples of 23 (moduli 23, 46, 69, 92), as shown in \autoref{fig:validcomparison}. Similar effects occur for multiples of 31, 39, and 47. Further visualizations can be found in Appendix~\ref{append:moduli}. Small moduli (1, 2, 4, 10, 12, 14, 15, 18) are learned within 100 epochs, while others exhibit stepwise grokking behavior. This moduli-specific learning suggests the model discovers mathematically meaningful functions. Therefore, we explore whether the underlying representations encapsulate relations between integers.

\paragraph{Visualizing the embedding space.} We perform PCA on token embeddings (tokens 1-100) before and after grokking to examine numerical patterns. We analyze embeddings by numeric value, lowest prime factor, parity, primality, divisor count, multiplicative order, Euler's totient function $\phi(n)$, primitive roots, residue classes modulo 5, and multiples of specific numbers (23, 31, 39). Before grokking, embeddings form spatially distinct clusters with weak structure for most number-theoretic properties. After grokking, embeddings become more centralized and compressed, though clear clustering by mathematical properties remains limited. The general centralization suggests structural reorganization during learning, but interpretable mathematical organization is not clearly evident. Additional PCA visualizations are provided in Appendix~\ref{sec:appendix_pca}.

\begin{figure}[t]
    \centering
    \includegraphics[width=0.9\linewidth]{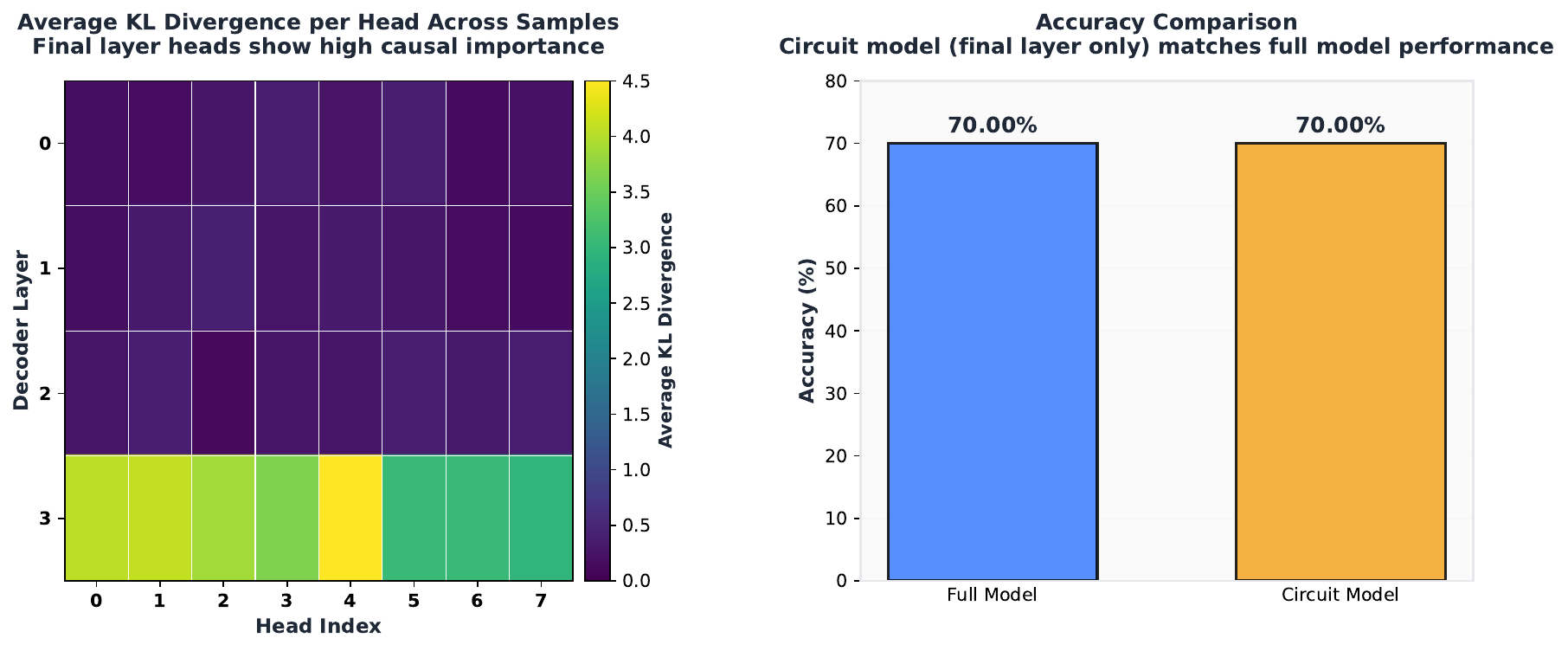}
\caption{\textit{Left}: KL divergence heatmap showing causal importance of each attention head across the 4-layer decoder. Warmer colors indicate higher KL divergence between clean and patched activations, reflecting greater causal impact on model predictions. Final-layer heads (layer 3) exhibit substantially higher KL divergence ($\sim$3--4.5) compared to earlier layers ($\sim$0.1--0.4), indicating that the circuit for regular exponentiation (when $c > a^b$) is concentrated in the final decoder layer. \textit{Right}: Accuracy comparison between the full model and the minimal circuit consisting only of final-layer attention heads. Both achieve 70\% accuracy on regular exponentiation tasks, demonstrating that earlier layers contribute negligibly to this computation and confirming functional specialization in the network architecture.}
    \label{fig:activation}
\end{figure}

\paragraph{Results of activation patching.} We use activation patching \citep{heimersheim2024useinterpretactivationpatching} to identify minimal circuits by replacing attention head activations with counterfactual inputs and measuring KL divergence. 
We find that regular exponentiation (when $c > a^b$) can be performed using only final-layer attention heads, achieving full model accuracy with a substantially smaller circuit, suggesting functional specialization where higher layers encode task-specific transformations.
Using 100 prompt-counterfactual pairs, circuit analysis reveals that final-layer attention heads alone achieve full model accuracy, with earlier layers having minimal causal impact (Figure \ref{fig:activation}). This suggests functional specialization where higher layers encode task-specific transformations.

\section{Discussion}

Our study extends transformer arithmetic learning to modular exponentiation. Reciprocal operand sampling achieves over 80\% test accuracy by resolving output space imbalance, echoing \citet{charton2024learning}'s findings that distributional choices significantly impact learning. The learning dynamics reveal moduli-specific grokking where accuracy surges coincide with solving sets of related moduli (e.g., multiples of 23), mirroring sieve-like learning in prior GCD work. 

PCA analysis shows embedding centralization post-grokking, though clear clustering by number-theoretic properties remains limited. The limited mathematical clustering in embeddings that we do observe suggests transformers may encode modular arithmetic through distributed representations rather than explicit symbolic groupings. The centralization post-grokking indicates structural reorganization, but further work with probing classifiers or feature attribution methods could better characterize what mathematical properties are captured. We also observe deterministic mispredictions (e.g., predicting 19 instead of 91) suggesting fallback mechanisms for rare inputs. Activation patching demonstrates that regular exponentiation uses only final-layer circuits, indicating functional specialization where transformers compartmentalize arithmetic operations.

\section{Conclusion}

We demonstrate that transformers can learn modular exponentiation with high accuracy using reciprocal sampling strategies. Key findings include stepwise grokking of related moduli, embedding space reorganization, and specialized circuits for arithmetic operations. 

\paragraph{Limitations.}Our findings are limited to compact transformers; scaling to larger architectures (e.g., 12+ layers, hundreds of attention heads) may reveal different circuit structures or learning dynamics. Future work should investigate whether specialized circuits persist or become distributed in larger models. We focus on synthetic data with moduli up to 100 and operands up to $10^6$. Real cryptographic applications use much larger bit-lengths (e.g., 2048-bit RSA). Our findings mainly demonstrate proof-of-concept for mechanistic understanding. Further, as our goal is mechanistic interpretability of transformers specifically, we do not compare against simpler sequence models (RNNs, LSTMs) or explicit algorithmic implementations. 

\section*{Acknowledgements}
We thank the reviewers for helpful feedback on limitations and scope. David Demitri Africa is supported by the Cambridge Trust and the Jardine Foundation. 

\bibliography{bib}

\appendix

\section{Appendix / supplemental material}

\subsection{Reciprocal operand distribution}\label{append:operanddistribution}
We would like to compute the probability distribution $G(n,0,M)$ for given $n \in \mathbb Z$ and $M$ using the reciprocal distribution $F(x,1,M+1)$. Note that the flooring simply implies squashing the probability mass of all values $n \le x < n+1$. Given the cumulative probability distribution $F(x,a,b)$ and the probability density function $f(x,a,b)$ of a reciprocal distribution defined on all reals, we have a distribution on the integers
\begin{equation*}
     H(n,a,b) = \int_{x \in [n, n+1]} f(x,a,b)~ dx = F(n+1,a,b) - F(n,a,b)
\end{equation*}
This can be case-split into
\begin{equation*}
H(n,a,b) = \begin{cases}0 & n < a  \mathrm{~or~} n \ge b\\
\frac{\ln(n+1) - \ln(n)}{\ln(b) - \ln(a)} & a \le n \le b - 1\\
\end{cases}
\end{equation*}
With $Y := X - 1$, $X \sim H$ as our shifted integer for our final distribution $P$, we substitute and simplify bounds and get for $Y = n'$
\begin{align*}
    P(Y=n',a,b) = H(n'+1,a,b) &= \begin{cases}0 & n' < a - 1 \mathrm{~or~} n' \ge b - 1\\
    \frac{\ln(n' + 2) - \ln(n'+1)}{\ln(b) - \ln(a)} & a - 1 \le n' \le b - 2\\
    \end{cases}
\end{align*}
Note that this still relies on the old bounds. To get the formulation used in the main part of the paper, set $a=1, b = M+2$.

\subsection{Additional PCA Embeddings Visualizations}
\label{sec:appendix_pca}
We present additional visualizations of the PCA embeddings presented in 4.4. These are 3D representations of the same 9 number-theoretic metrics, and in addition a visualization by the multiples of 23 for which we observed significant performance increases, outlined in 4.3. As with the other metrics, performing PCA on multiples did not display any notable clusterings before grokking, with a centralization of embeddings emerging post-grokking.

\begin{figure}[htbp]
    \centering

    \begin{subfigure}[b]{0.3\textwidth}
        \includegraphics[width=4.2cm]{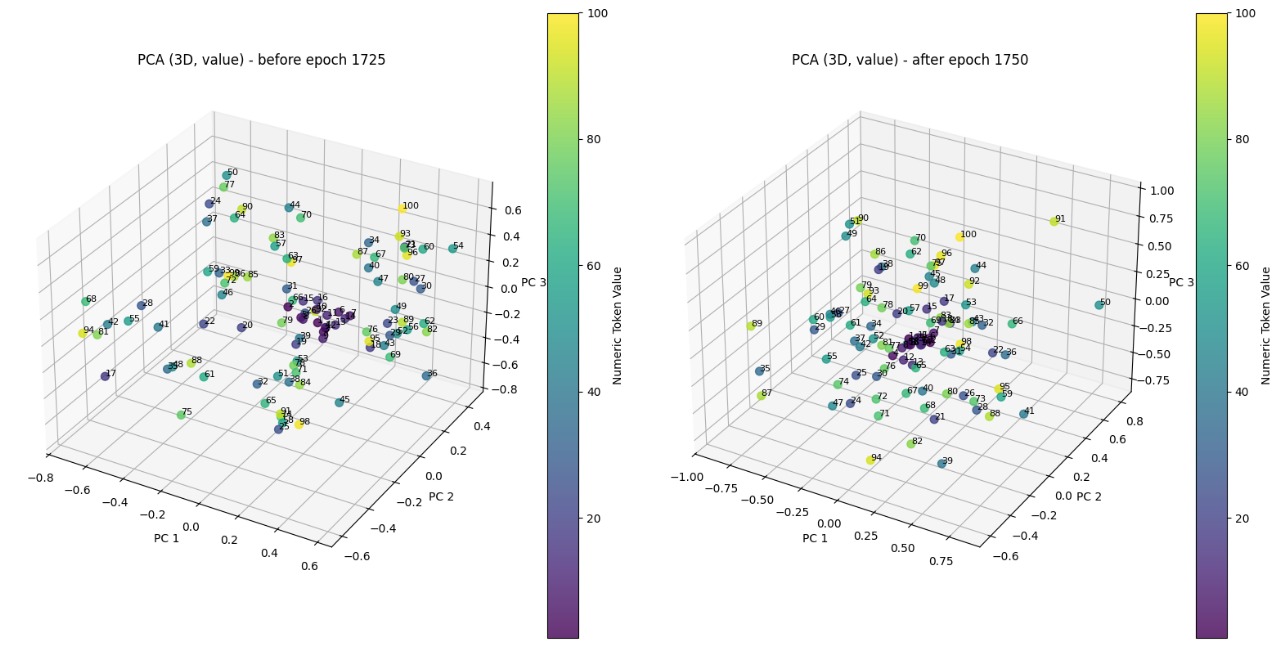}
        \caption{Value (3D)}
    \end{subfigure}
    \begin{subfigure}[b]{0.3\textwidth}
        \includegraphics[width=4.2cm]{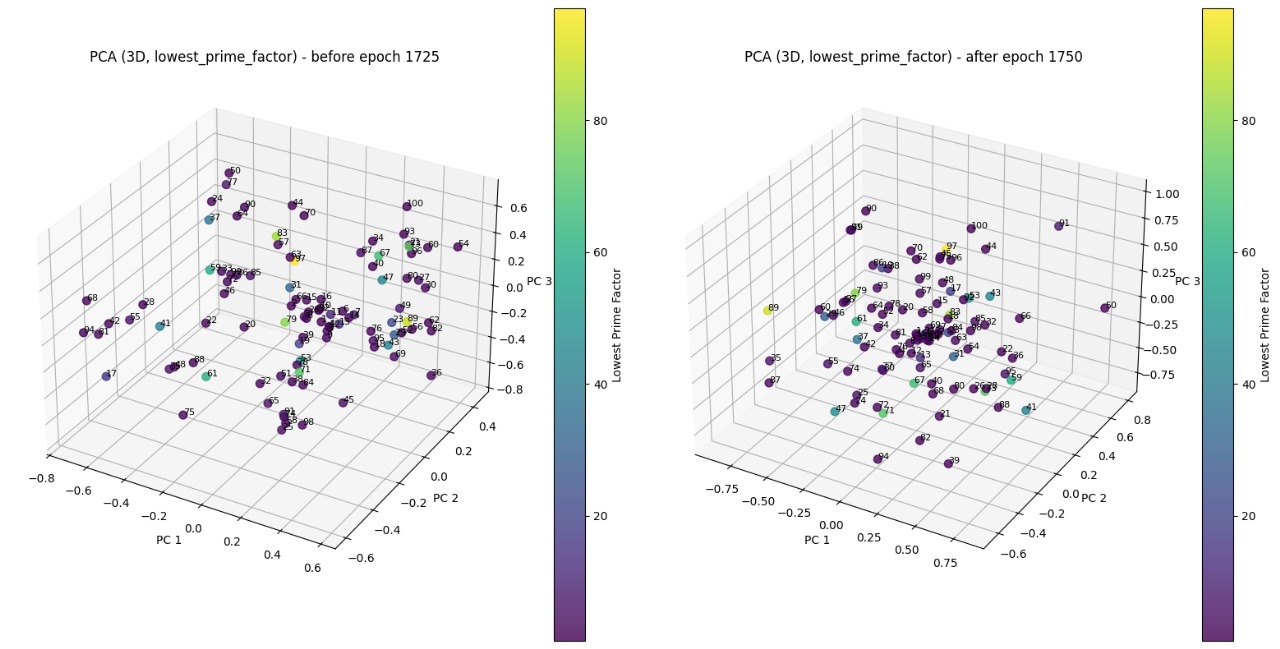}
        \caption{Lowest Prime Factor (3D)}
    \end{subfigure}
    \begin{subfigure}[b]{0.3\textwidth}
        \includegraphics[width=4.2cm]{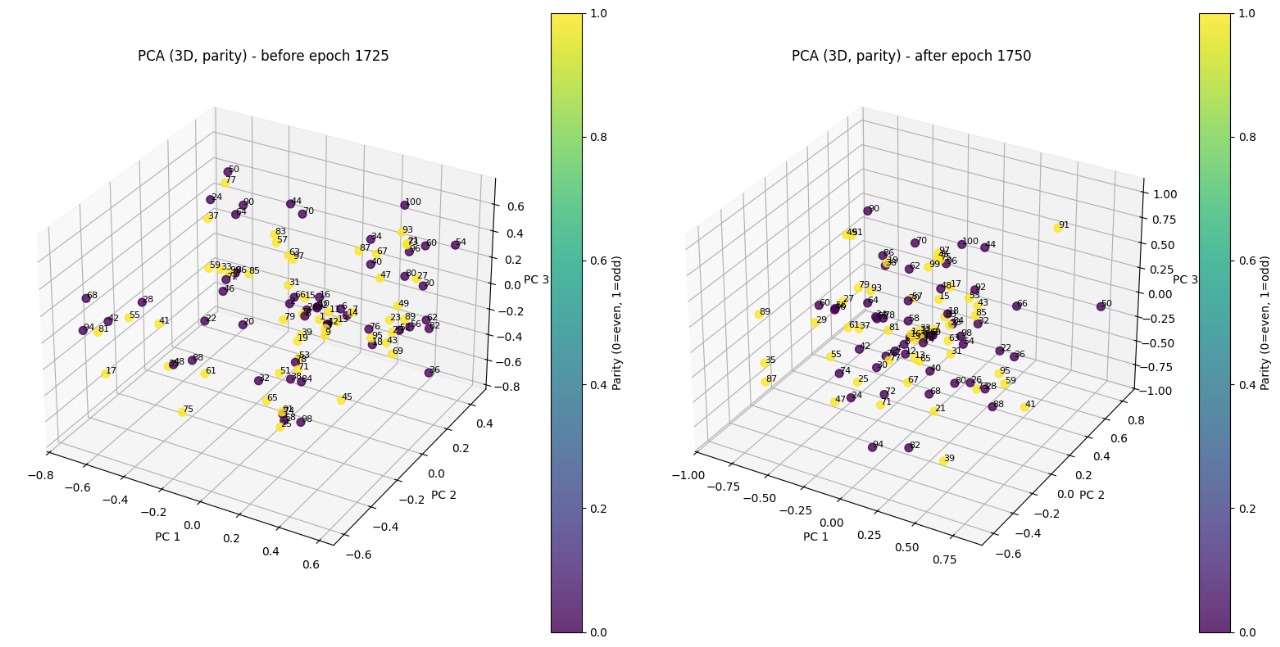}
        \caption{Parity (3D)}
    \end{subfigure}

    \begin{subfigure}[b]{0.3\textwidth}
        \includegraphics[width=4.2cm]{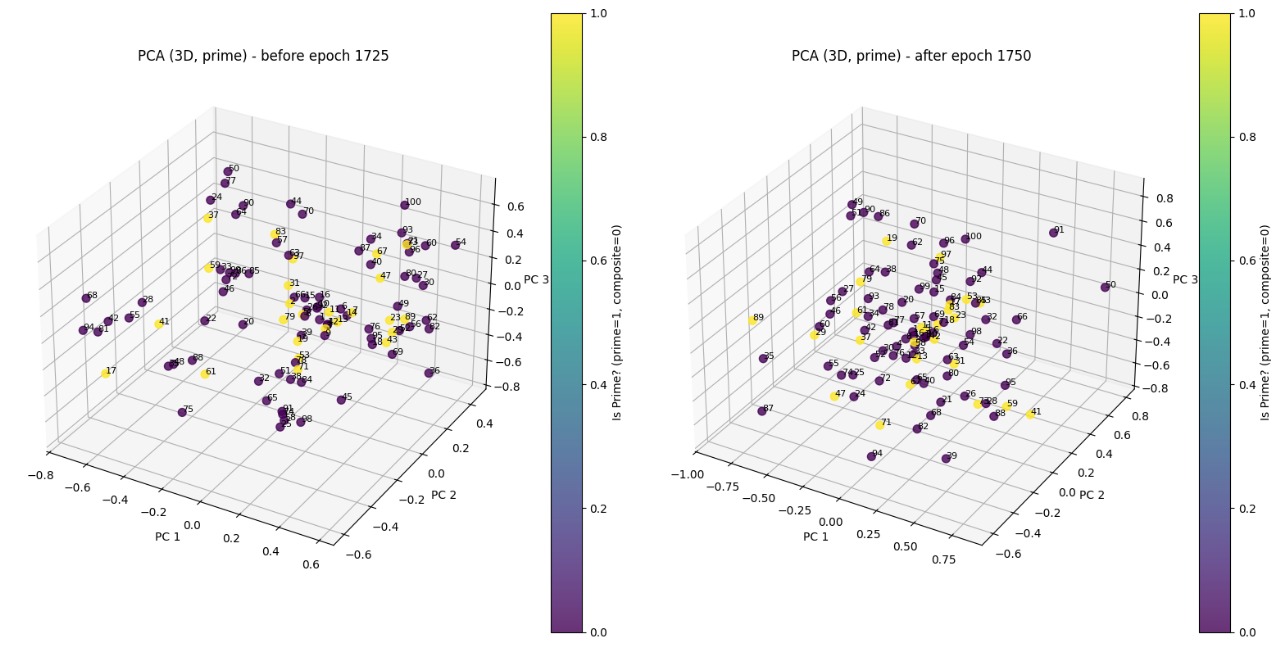}
        \caption{Prime (3D)}
    \end{subfigure}
    \begin{subfigure}[b]{0.3\textwidth}
        \includegraphics[width=4.2cm]{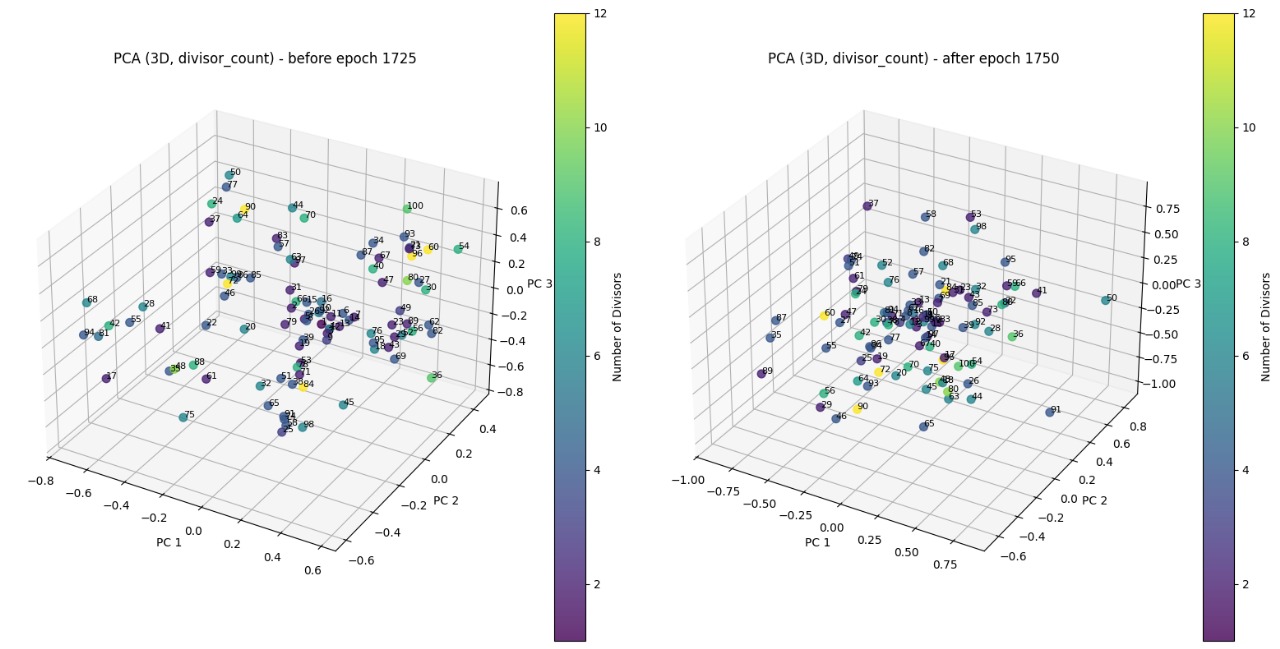}
        \caption{Divisor Count (3D)}
    \end{subfigure}
    \begin{subfigure}[b]{0.3\textwidth}
        \includegraphics[width=4.2cm]{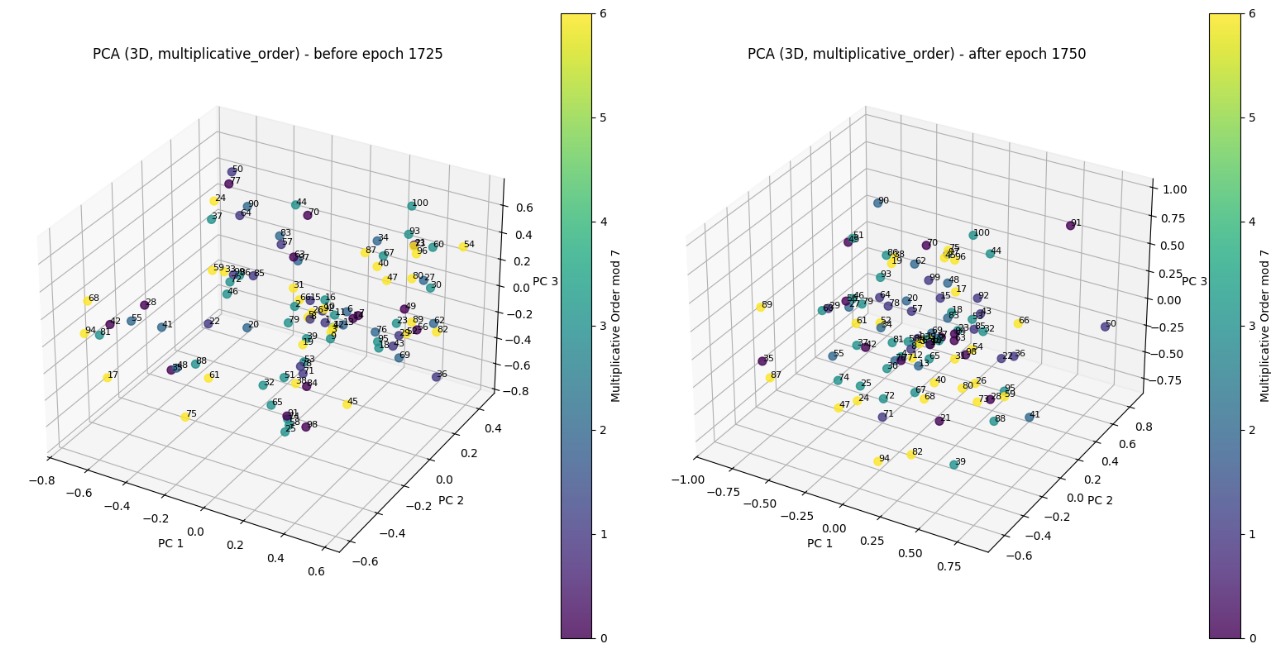}
        \caption{Multiplicative Order (3D)}
    \end{subfigure}

    \begin{subfigure}[b]{0.3\textwidth}
        \includegraphics[width=4.2cm]{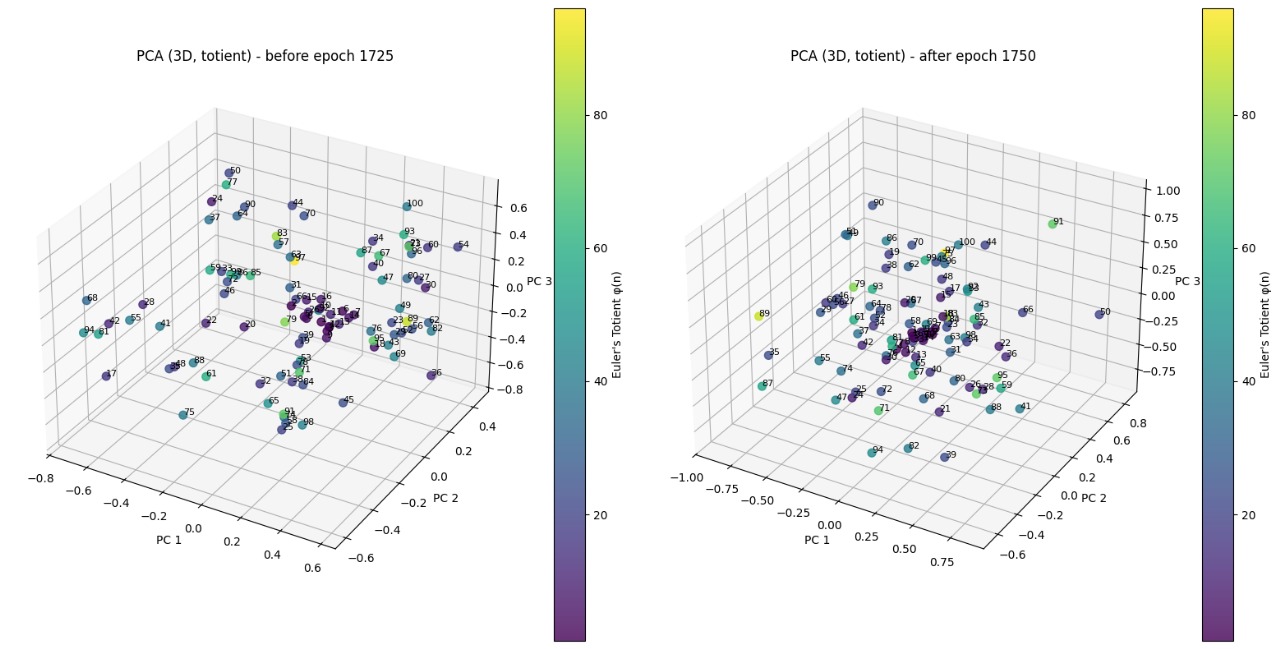}
        \caption{Euler’s Totient $\phi(n)$ (3D)}
    \end{subfigure}
    \begin{subfigure}[b]{0.3\textwidth}
        \includegraphics[width=4.2cm]{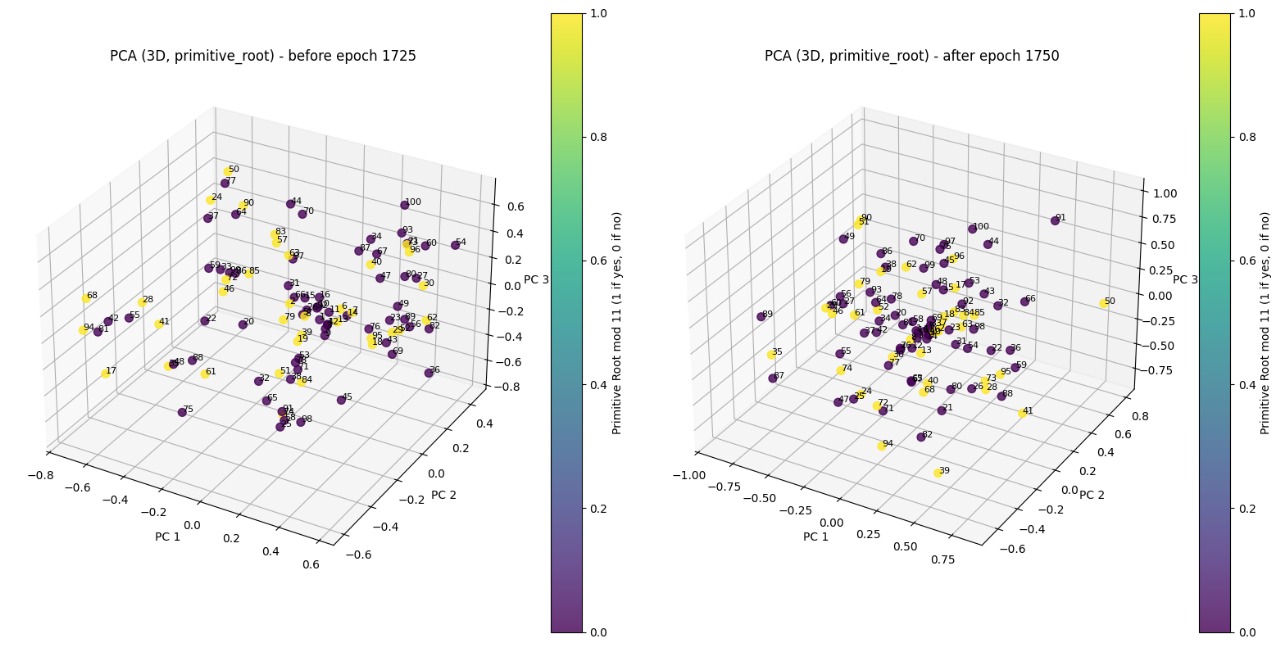}
        \caption{Primitive Root (3D)}
    \end{subfigure}
    \begin{subfigure}[b]{0.3\textwidth}
        \includegraphics[width=4.2cm]{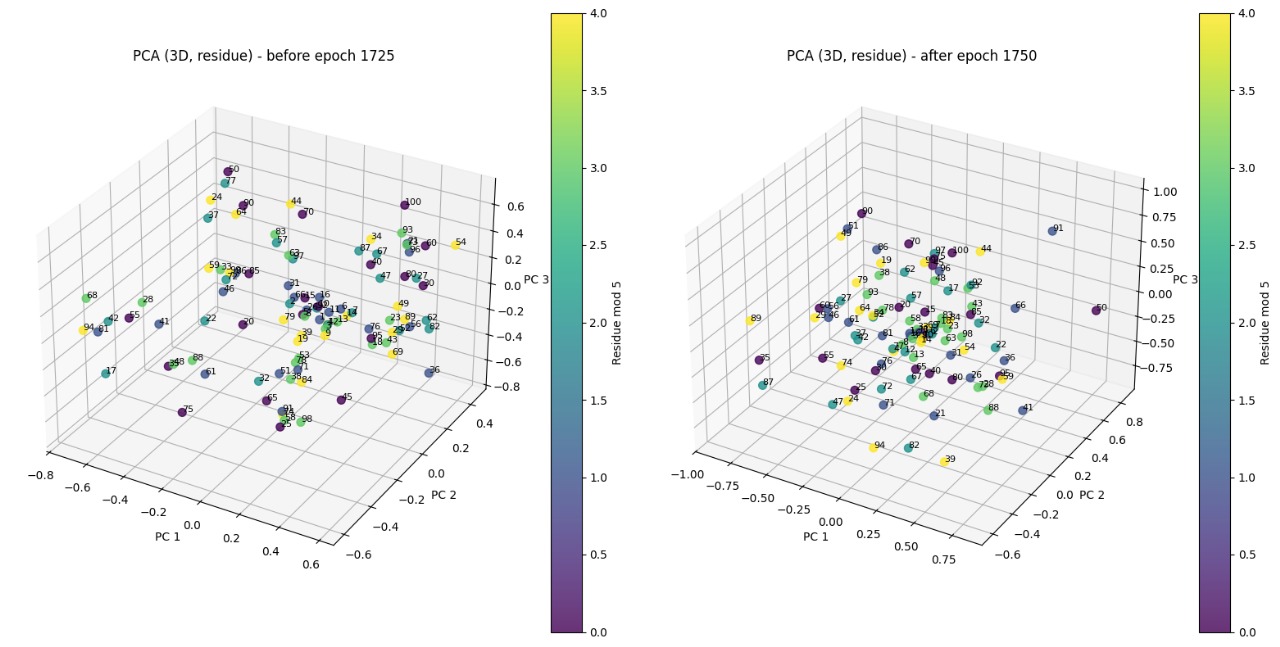}
        \caption{Residue mod 5 (3D)}
    \end{subfigure}

    \begin{subfigure}[b]{0.3\textwidth}
        \includegraphics[width=4.2cm]{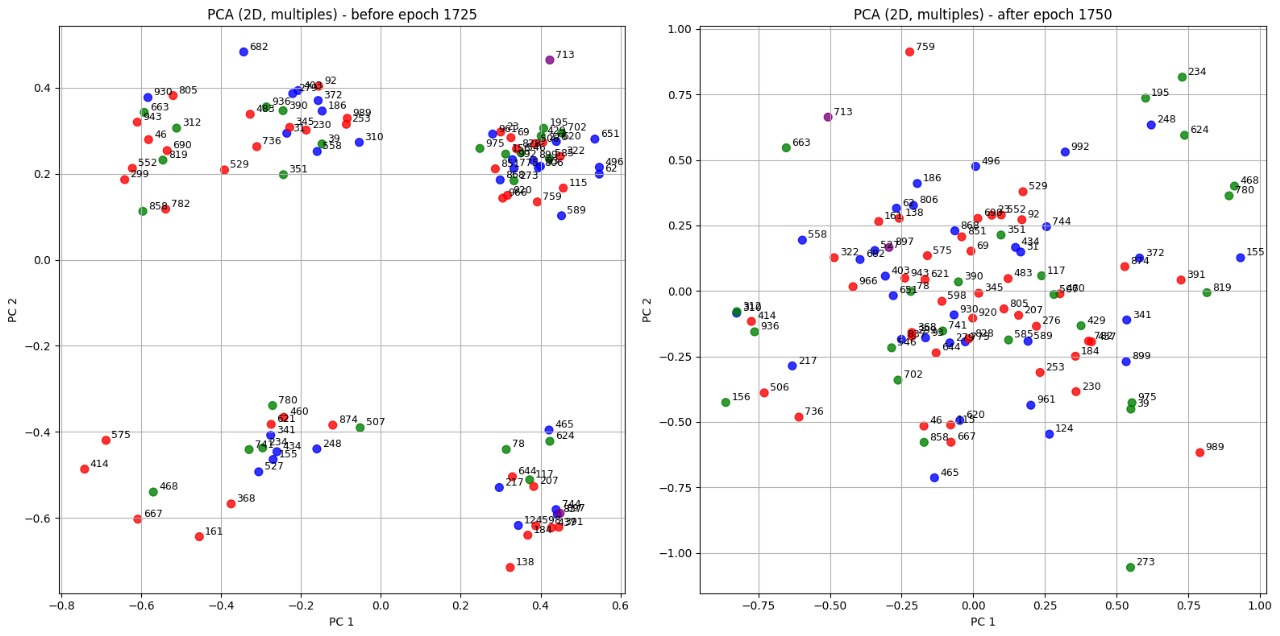}
        \caption{Multiples (2D)}
    \end{subfigure}

    \caption{PCA 3D projections of token embeddings, colored by number-theoretic properties, before and after grokking. Bottom row shows multiples in 2D.}
    \label{fig:pca_3d_main}
\end{figure}

\subsection{Additional moduli}\label{append:moduli}
We present some additional charts showing the learning dynamics of various moduli. The second to last number separated by \_ encodes the modulus.

\begin{figure}[htbp]
    \centering

    \begin{minipage}[b]{0.30\linewidth}
        \includegraphics[width=\linewidth]{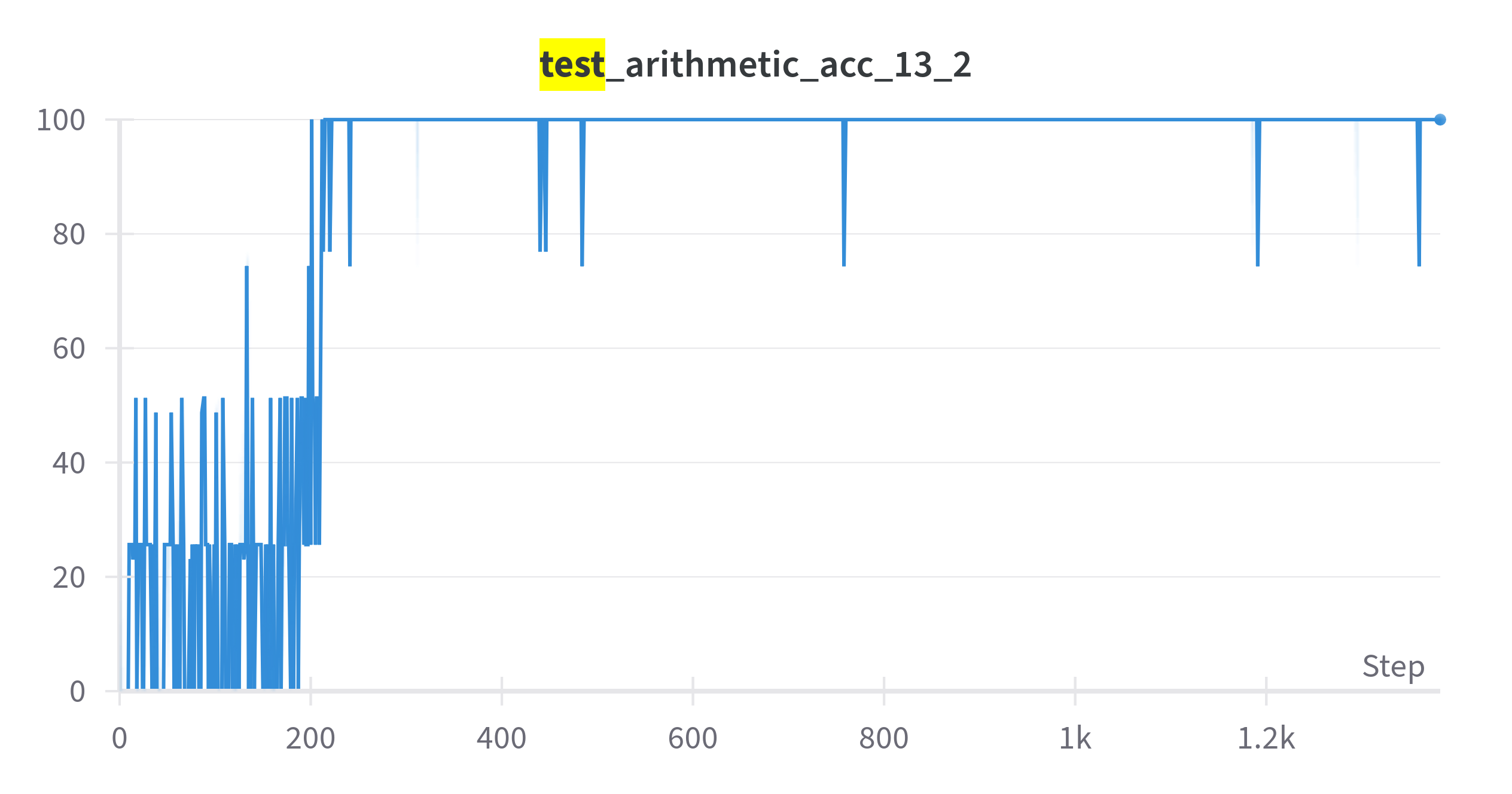}
    \end{minipage} \hfill
    \begin{minipage}[b]{0.30\linewidth}
        \includegraphics[width=\linewidth]{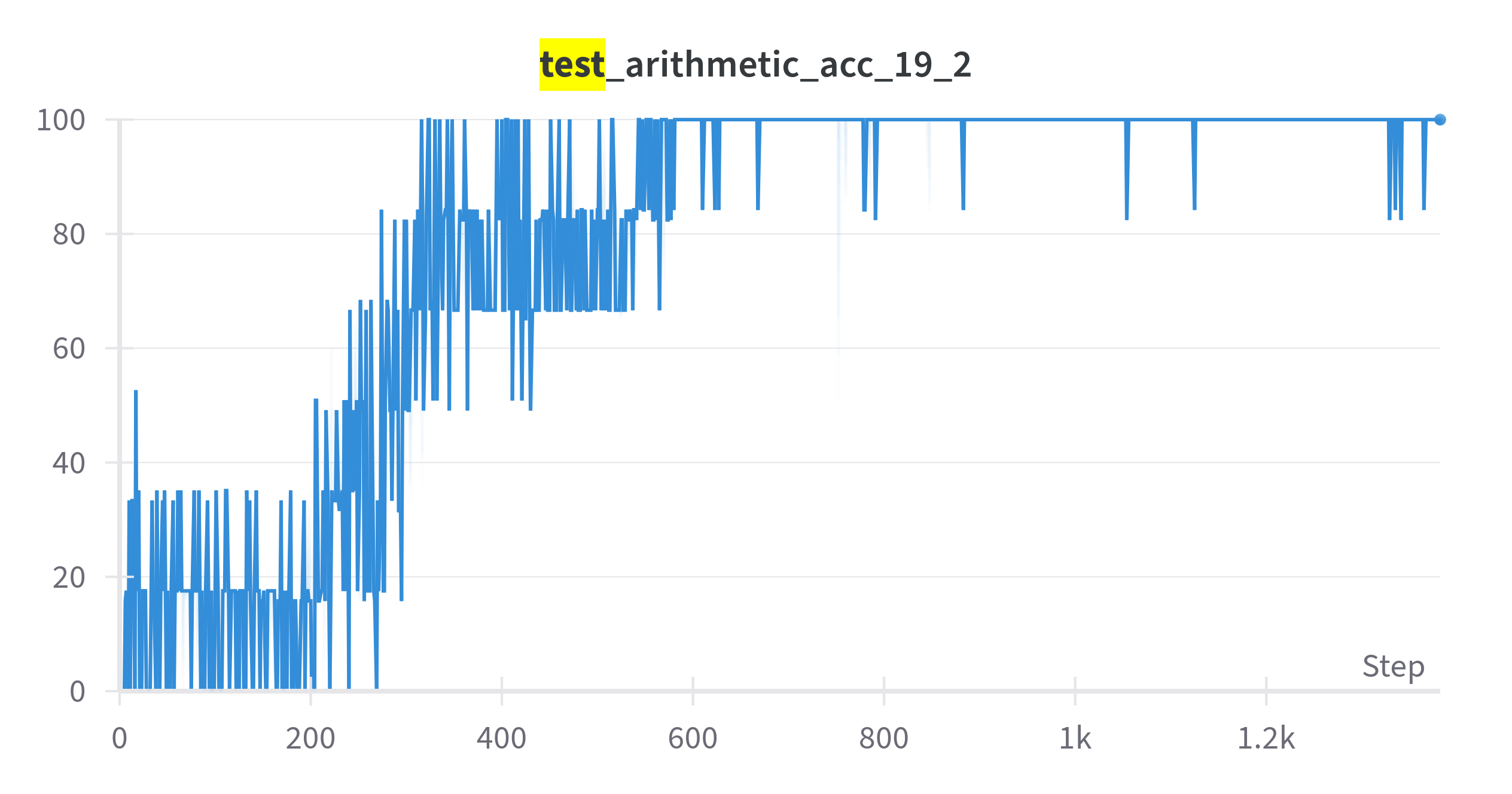}
    \end{minipage} \hfill
    \begin{minipage}[b]{0.30\linewidth}
        \includegraphics[width=\linewidth]{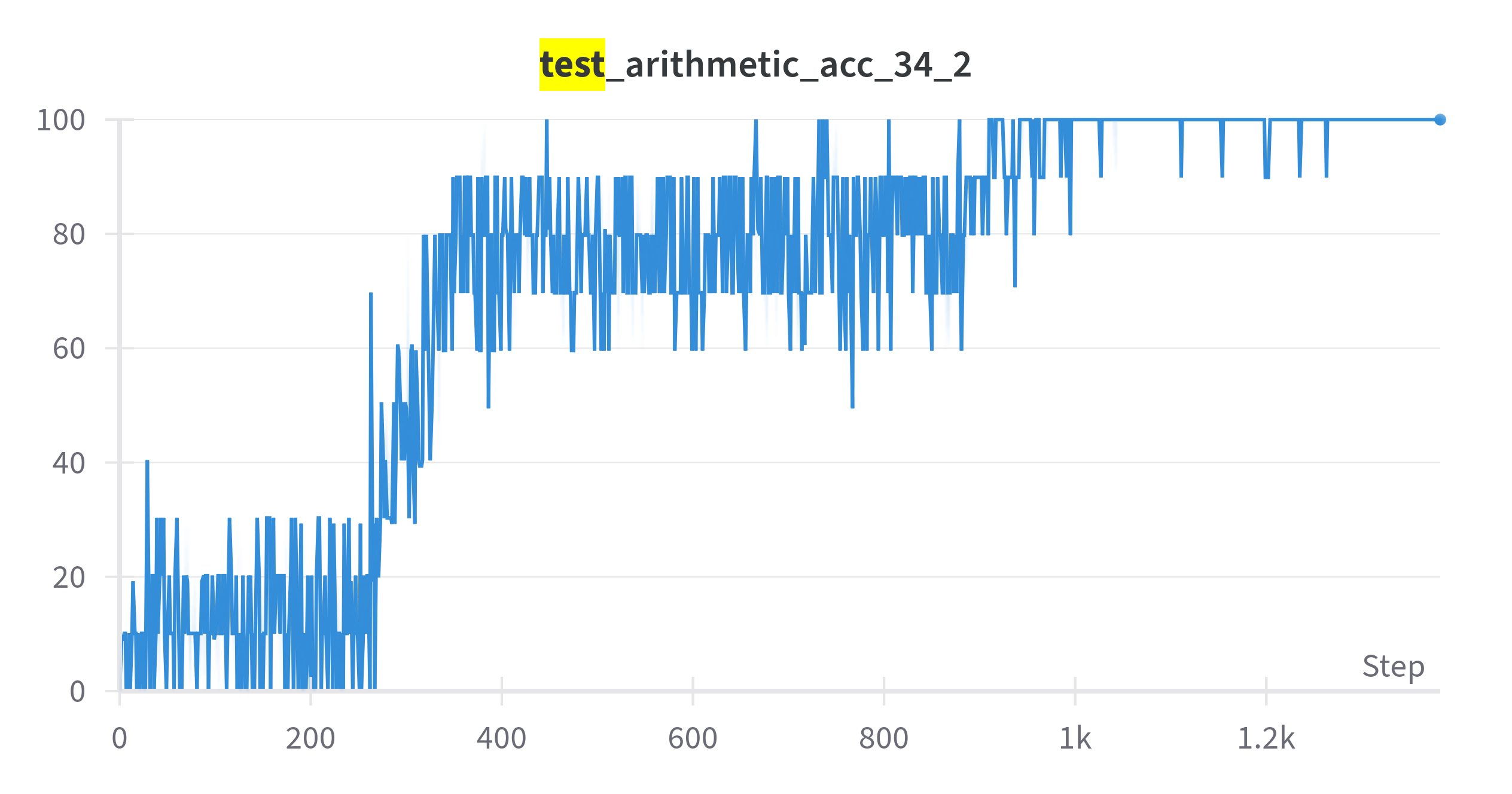}
    \end{minipage}

    \vspace{0.5em}

    \begin{minipage}[b]{0.30\linewidth}
        \includegraphics[width=\linewidth]{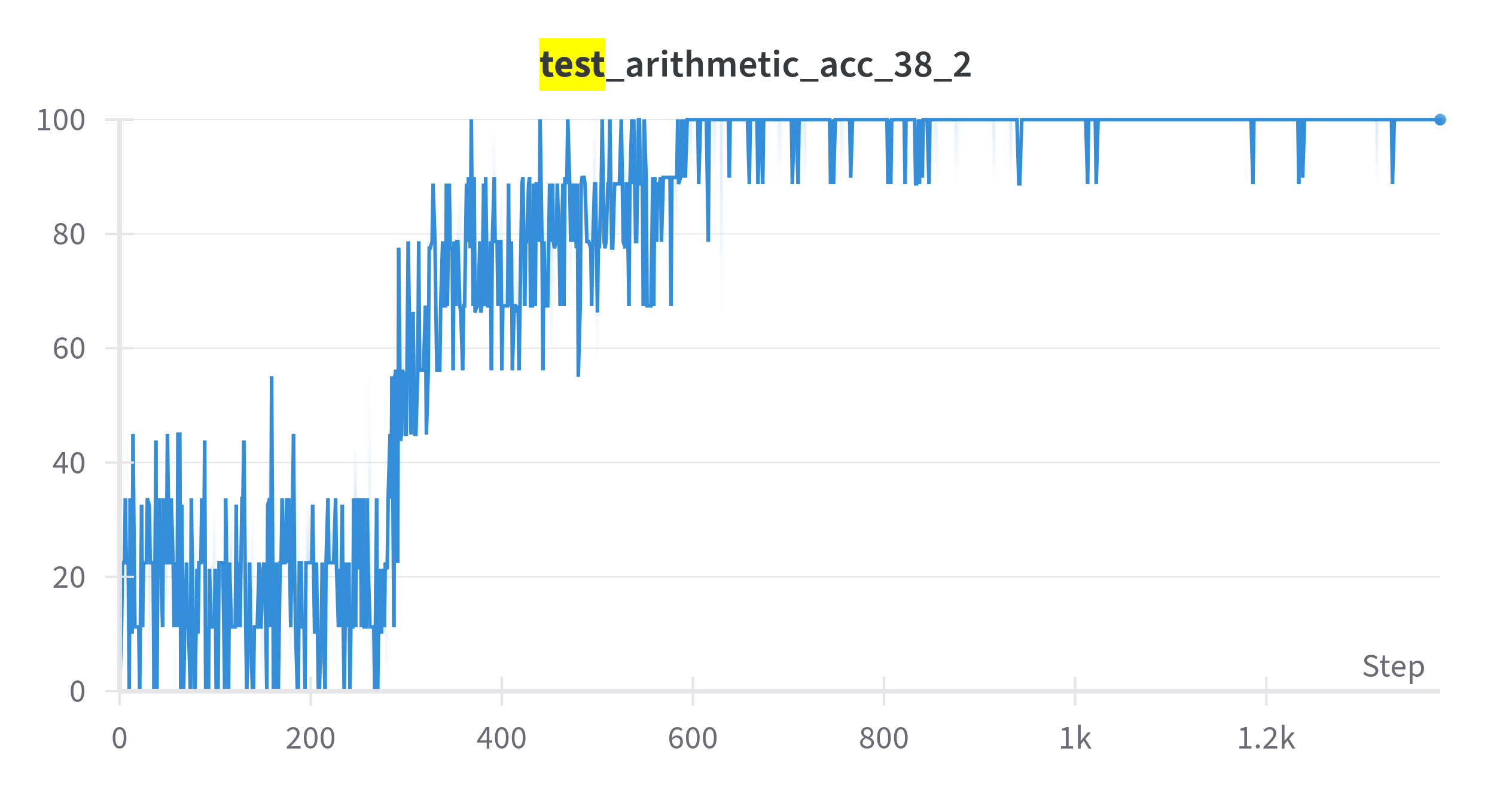}
    \end{minipage} \hfill
    \begin{minipage}[b]{0.30\linewidth}
        \includegraphics[width=\linewidth]{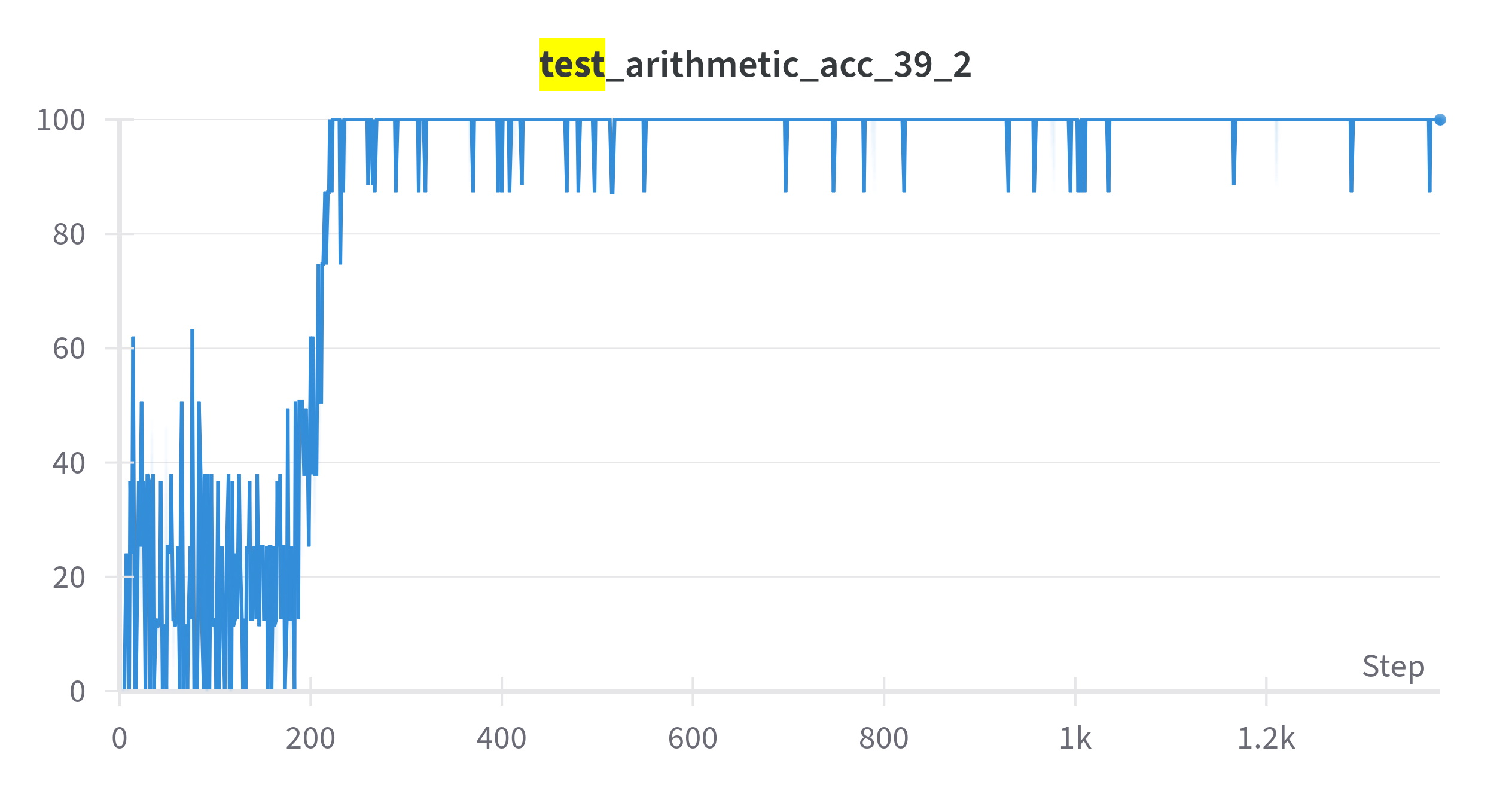}
    \end{minipage} \hfill
    \begin{minipage}[b]{0.30\linewidth}
        \includegraphics[width=\linewidth]{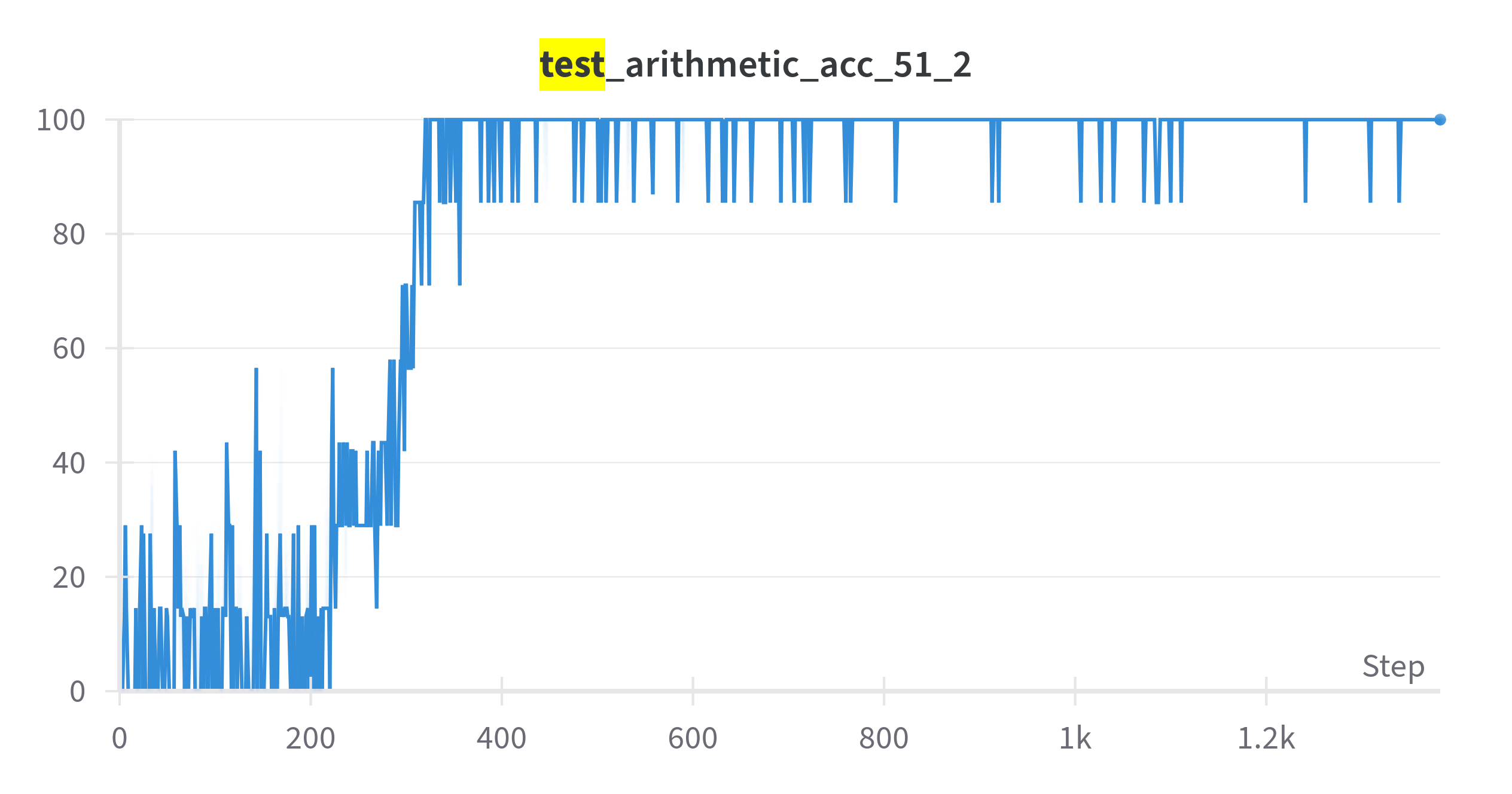}
    \end{minipage}

    \vspace{0.5em}

    \begin{minipage}[b]{0.30\linewidth}
        \includegraphics[width=\linewidth]{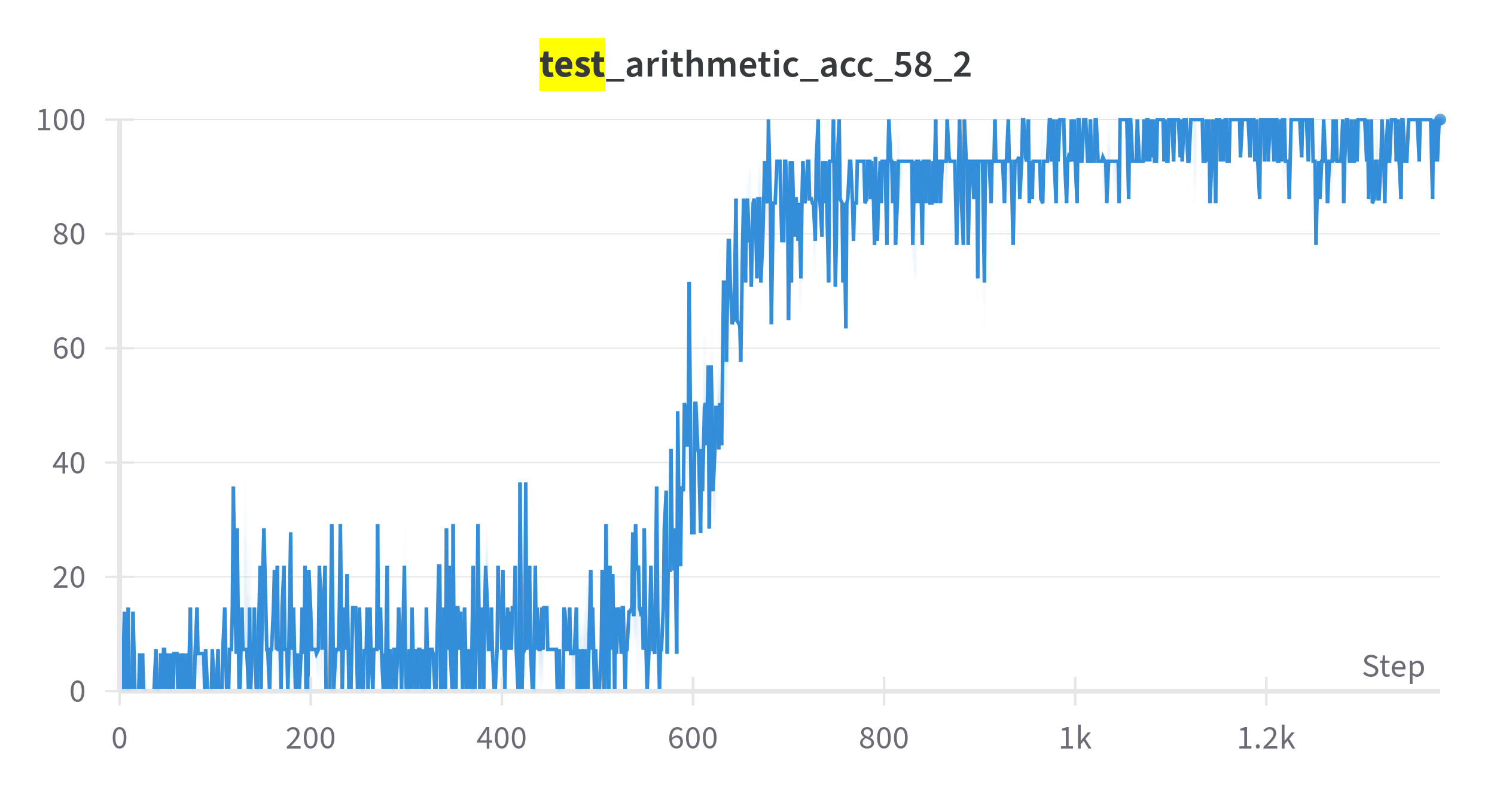}
    \end{minipage} \hfill
    \begin{minipage}[b]{0.30\linewidth}
        \includegraphics[width=\linewidth]{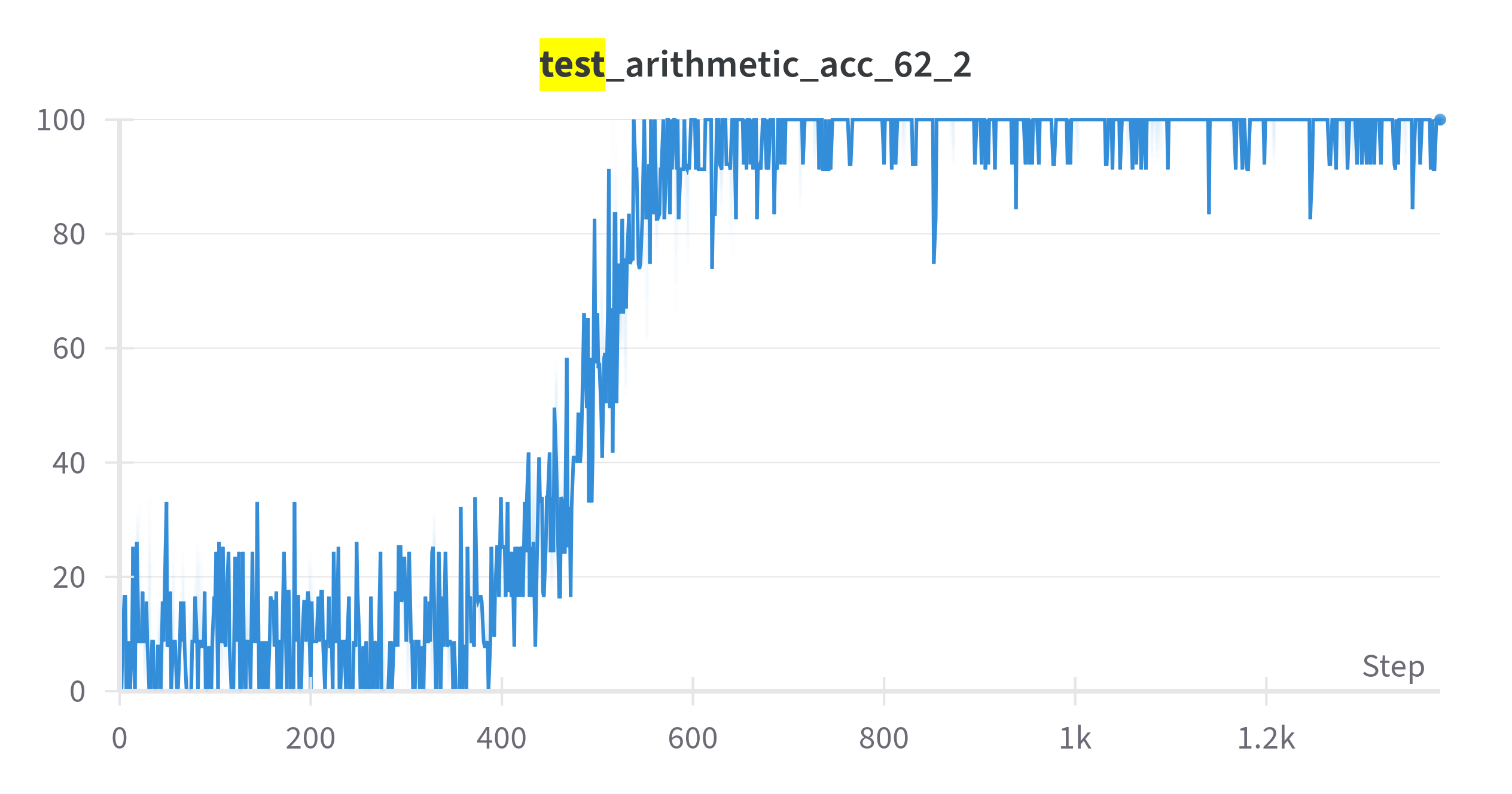}
    \end{minipage} \hfill
    \begin{minipage}[b]{0.30\linewidth}
        \includegraphics[width=\linewidth]{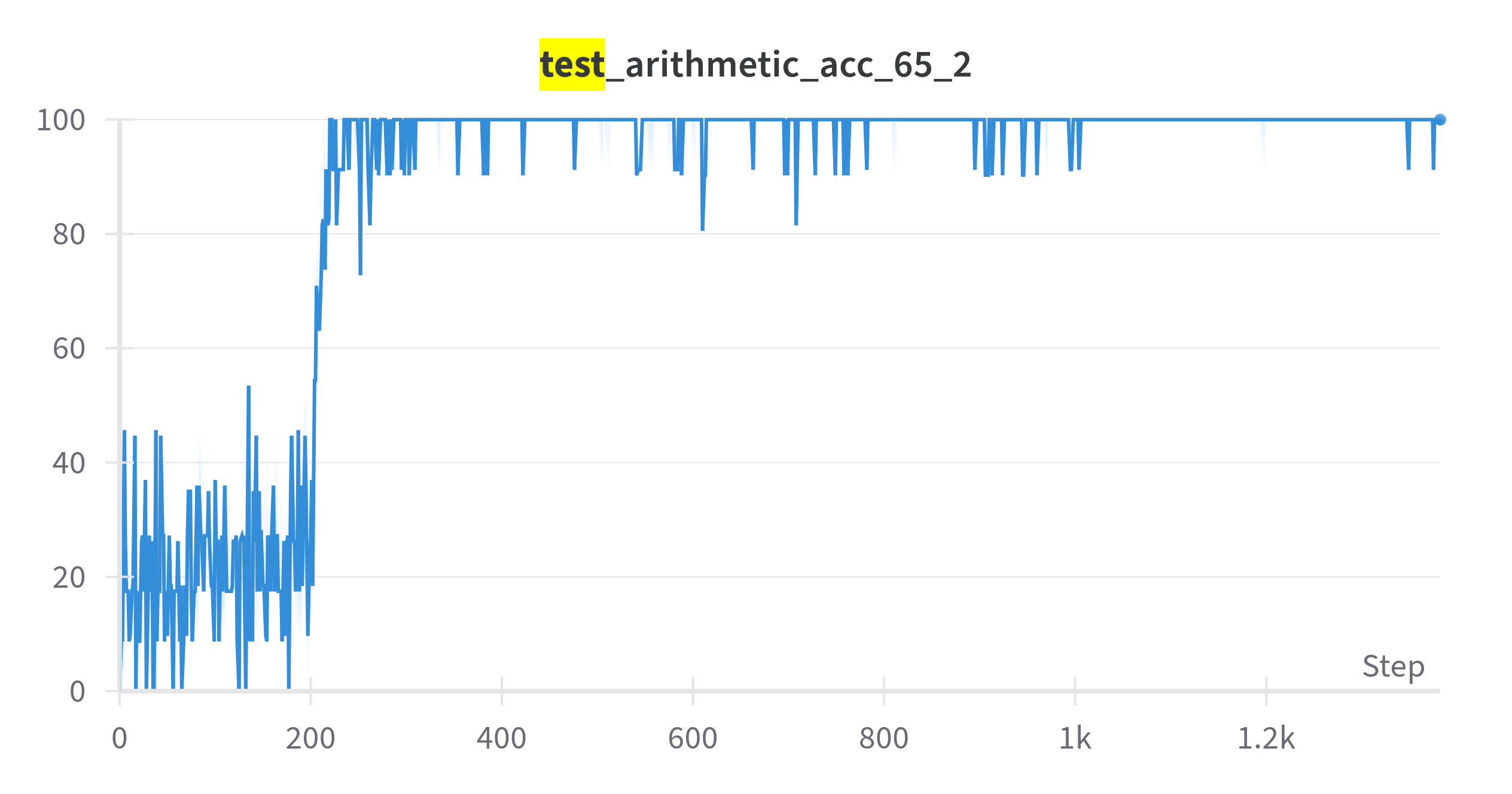}
    \end{minipage}

    \vspace{0.5em}

    \begin{minipage}[b]{0.30\linewidth}
        \includegraphics[width=\linewidth]{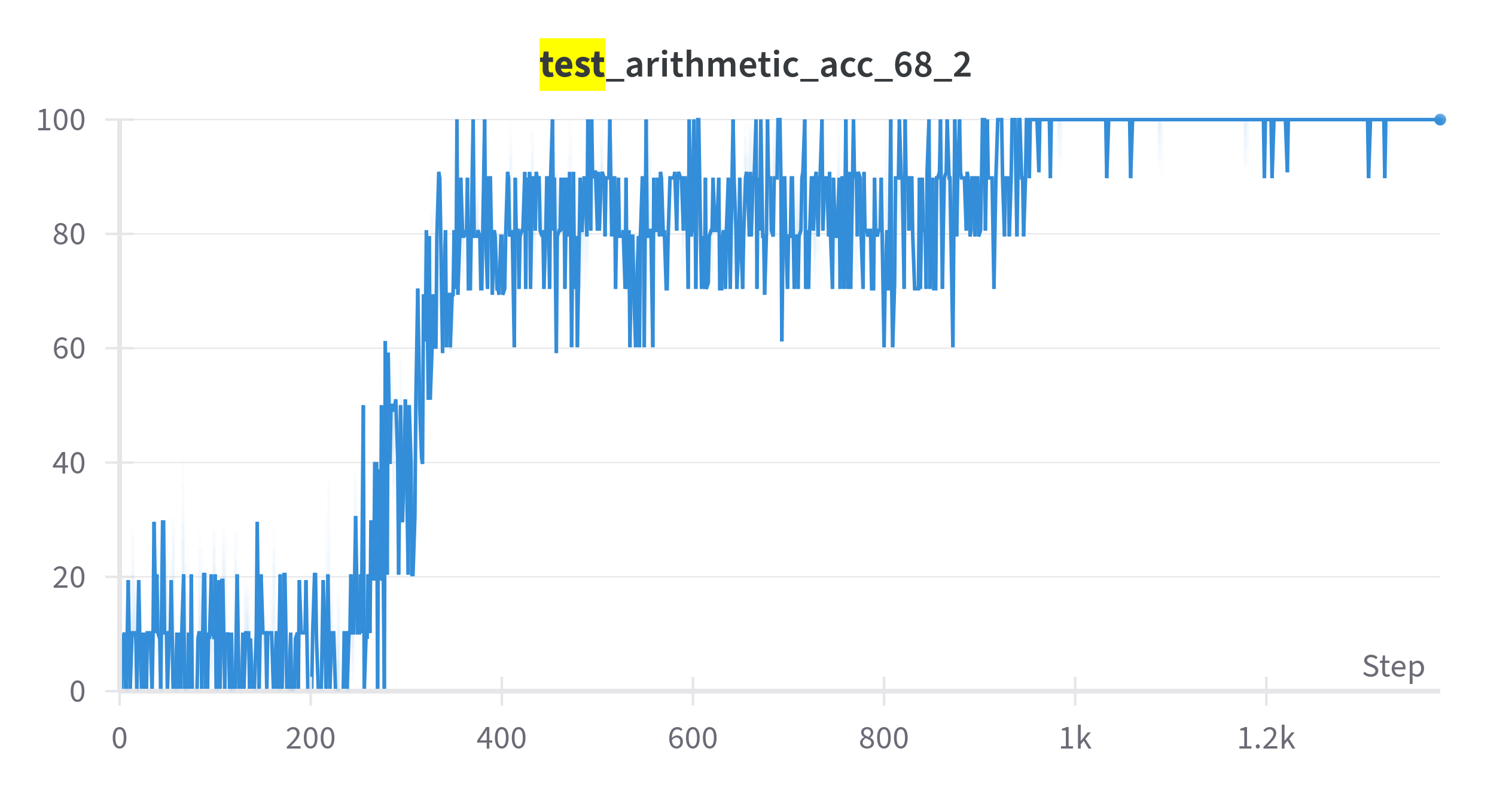}
    \end{minipage} \hfill
    \begin{minipage}[b]{0.30\linewidth}
        \includegraphics[width=\linewidth]{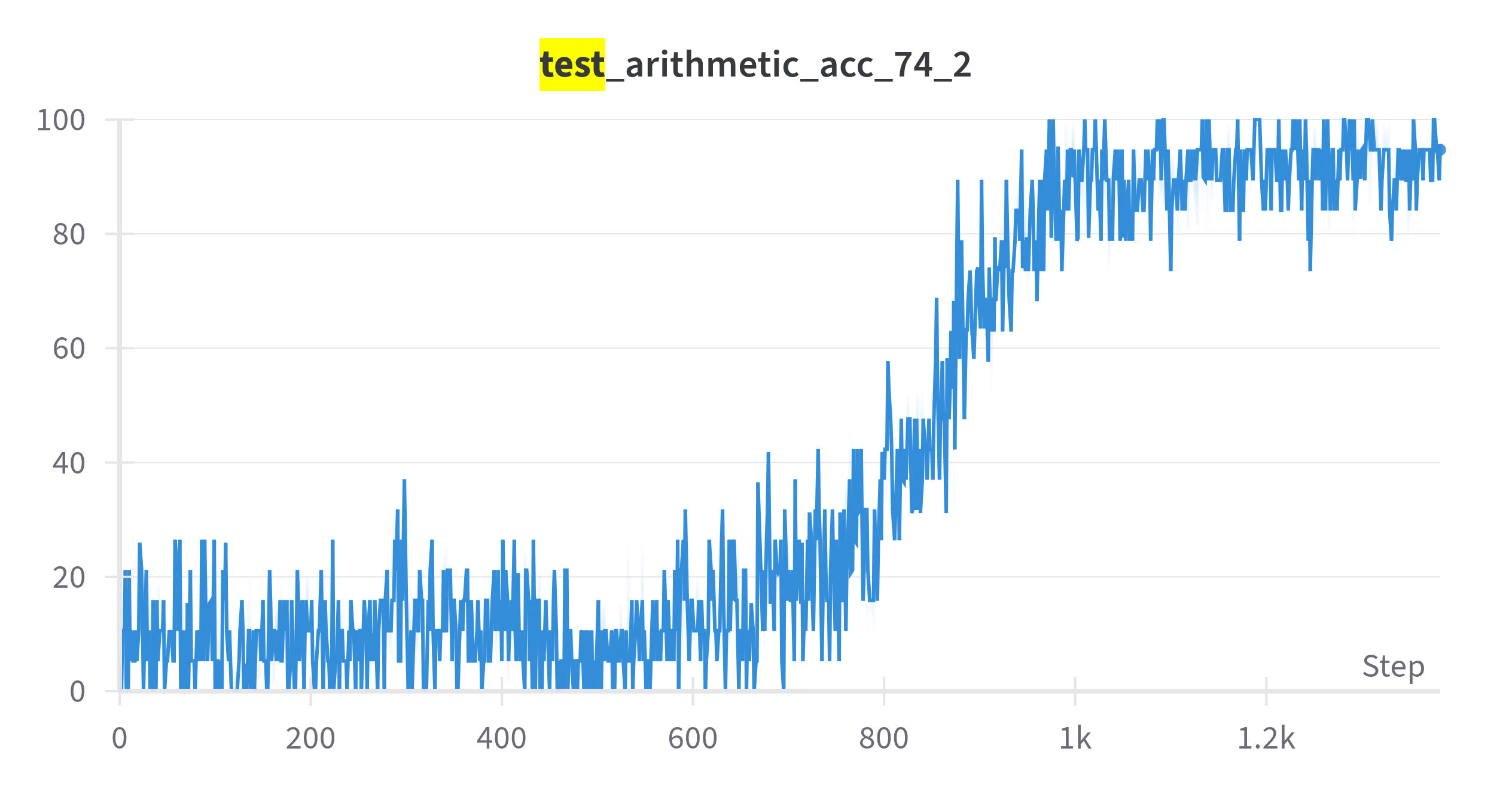}
    \end{minipage} \hfill
    \begin{minipage}[b]{0.30\linewidth}
        \includegraphics[width=\linewidth]{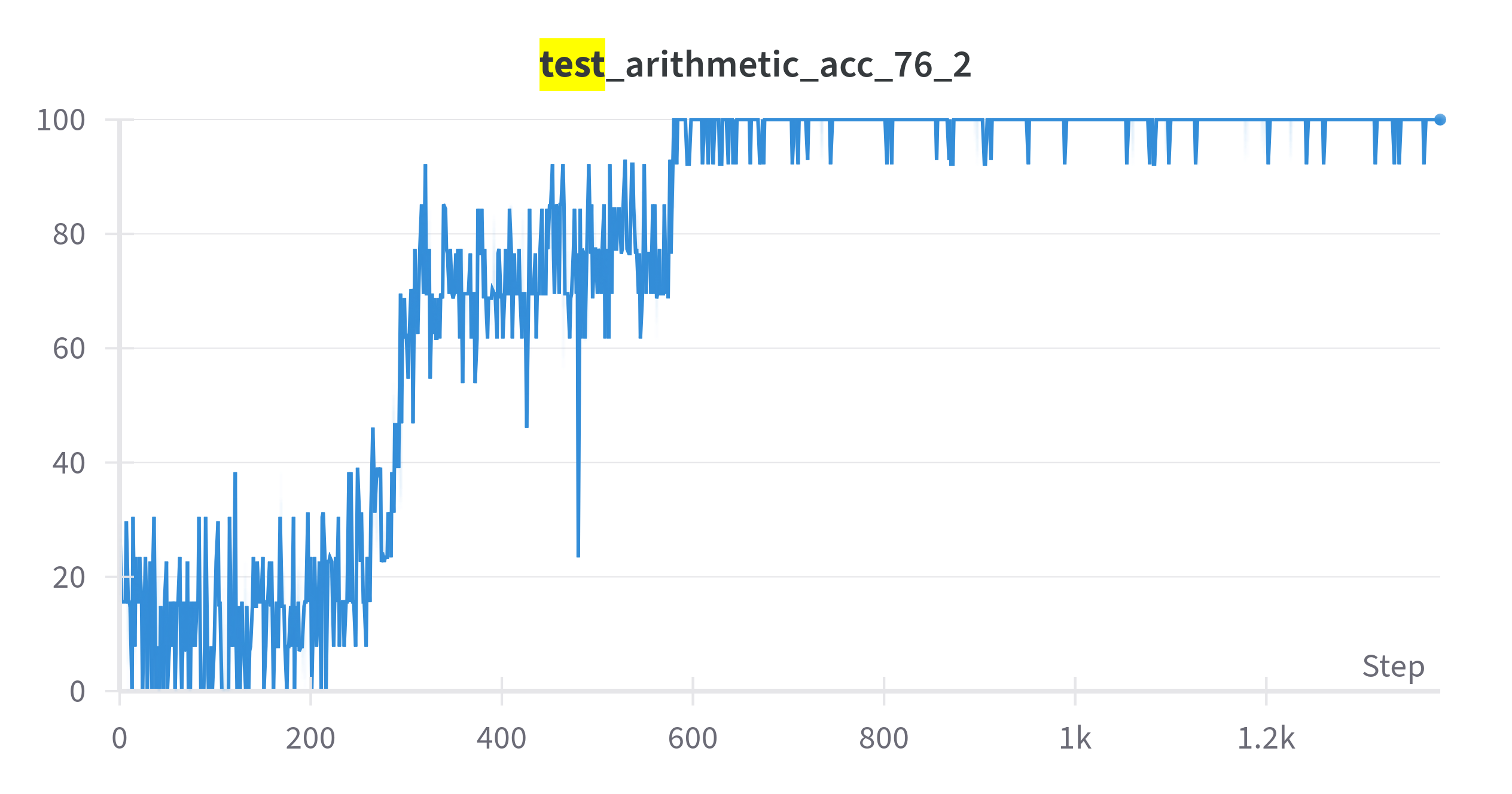}
    \end{minipage}

    \vspace{0.5em}

    \begin{minipage}[b]{0.30\linewidth}
        \includegraphics[width=\linewidth]{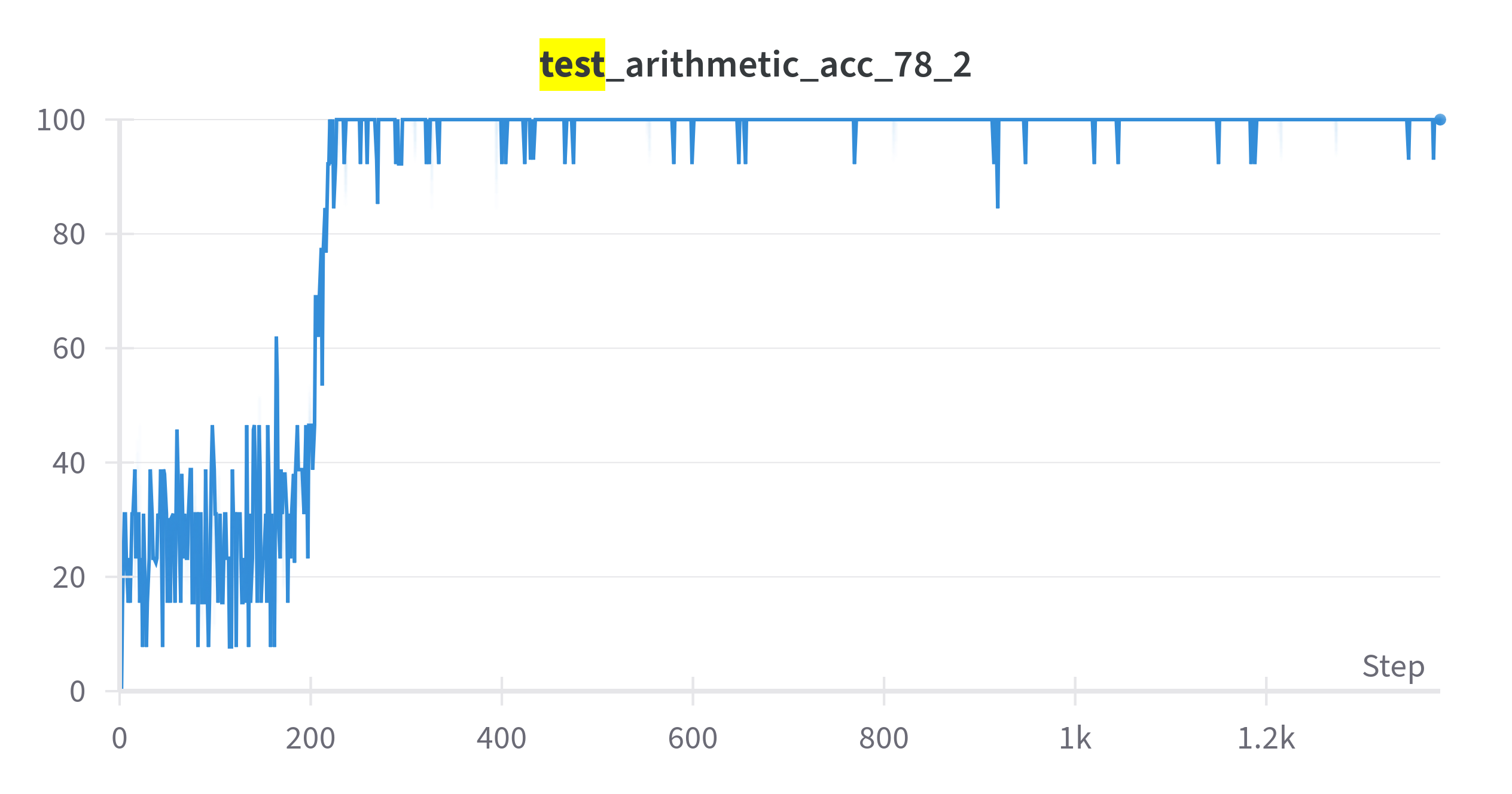}
    \end{minipage} \hfill
    \begin{minipage}[b]{0.30\linewidth}
        \includegraphics[width=\linewidth]{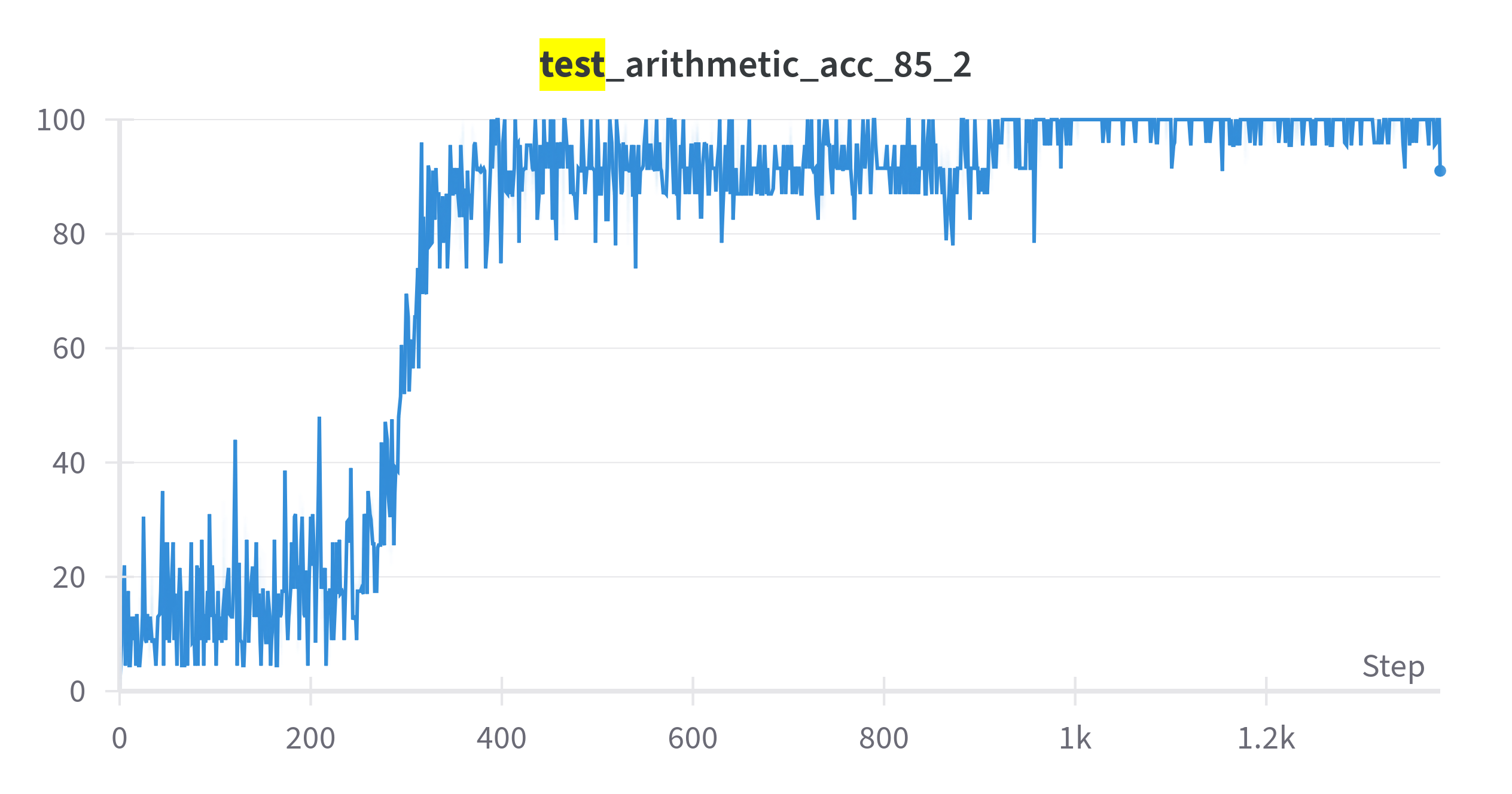}
    \end{minipage} \hfill
    \begin{minipage}[b]{0.30\linewidth}
        \includegraphics[width=\linewidth]{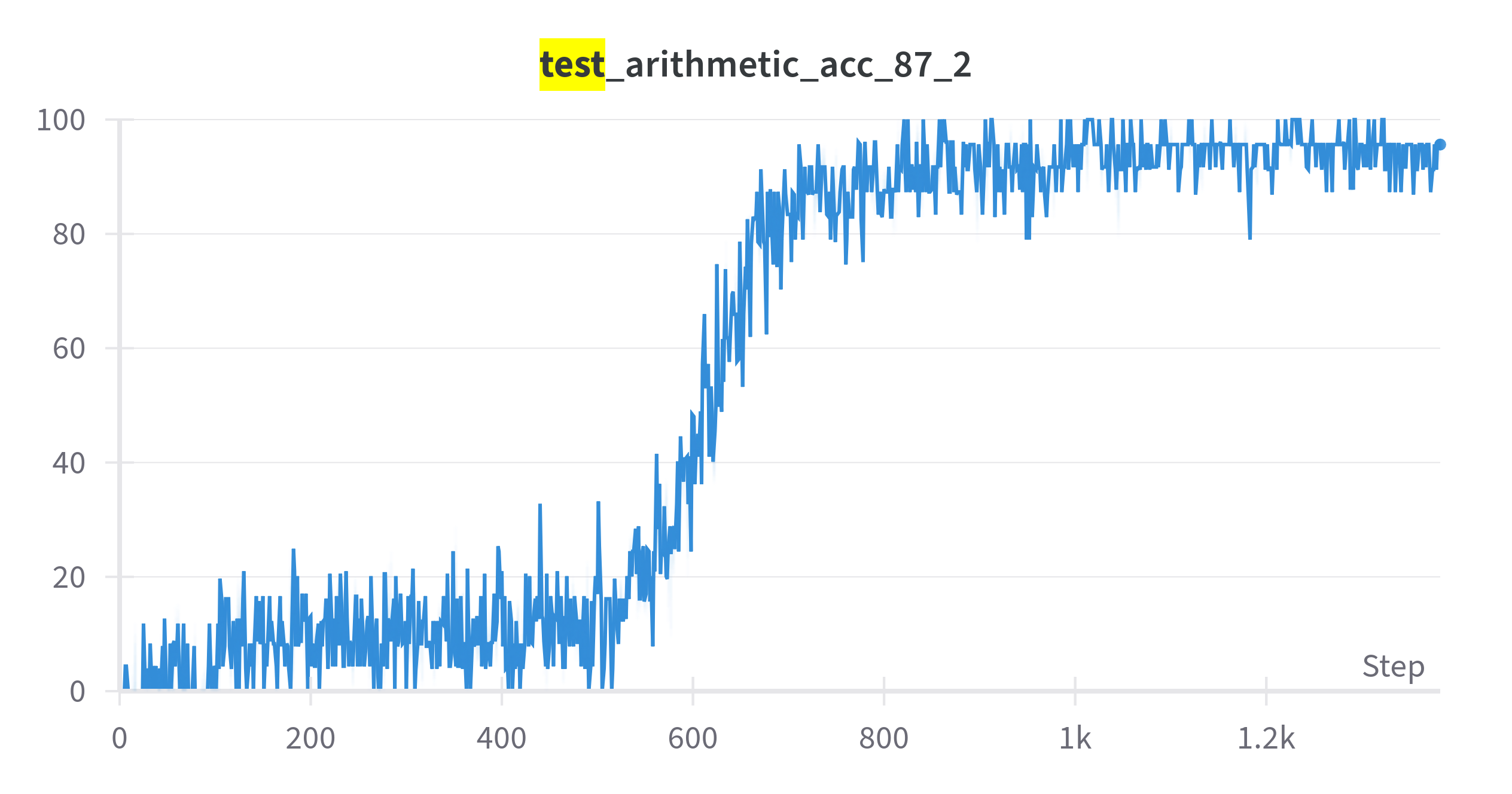}
    \end{minipage}

    \vspace{0.5em}

    \begin{minipage}[b]{0.30\linewidth}
        \includegraphics[width=\linewidth]{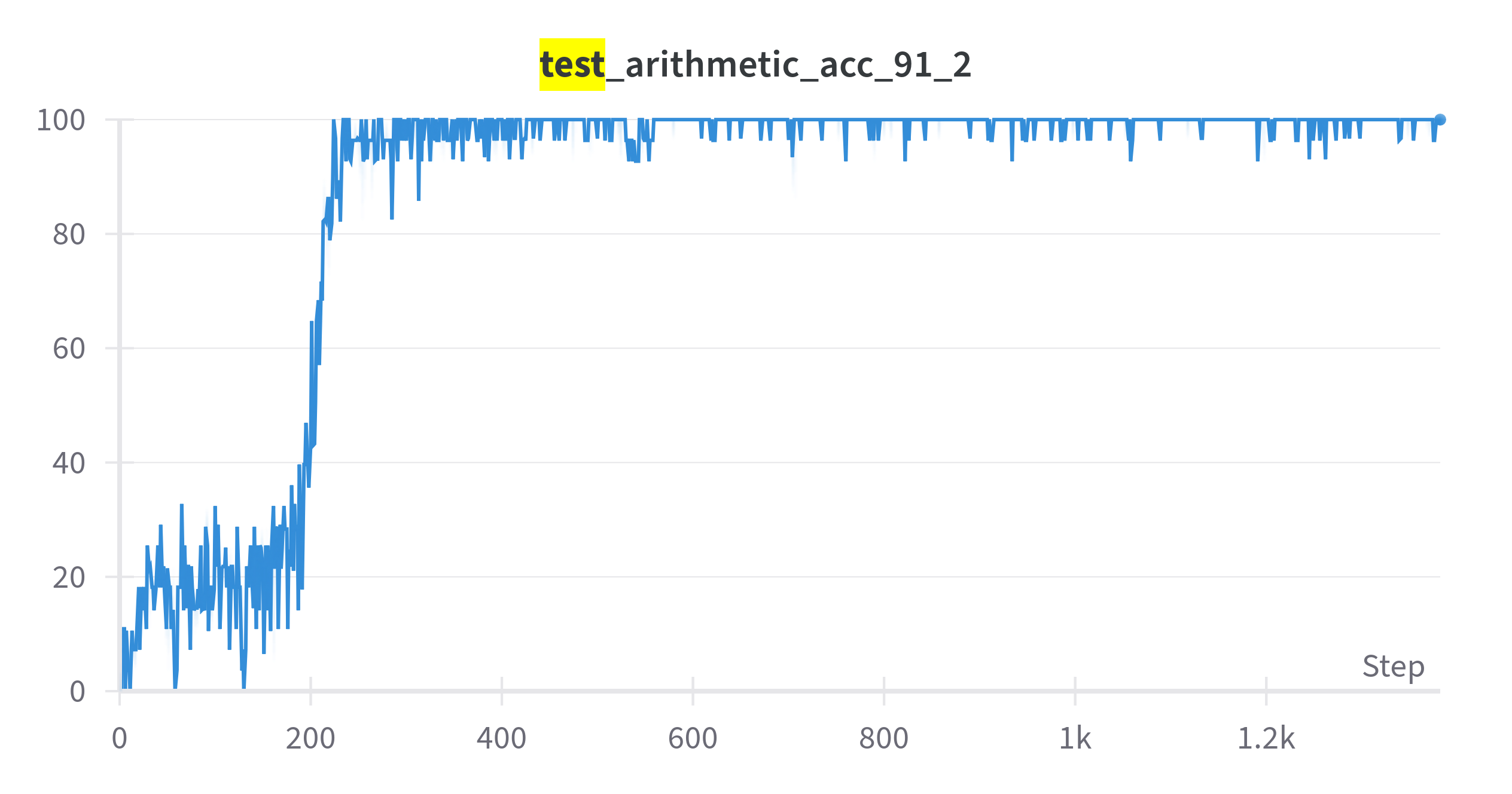}
    \end{minipage} \hfill
    \begin{minipage}[b]{0.30\linewidth}
        \includegraphics[width=\linewidth]{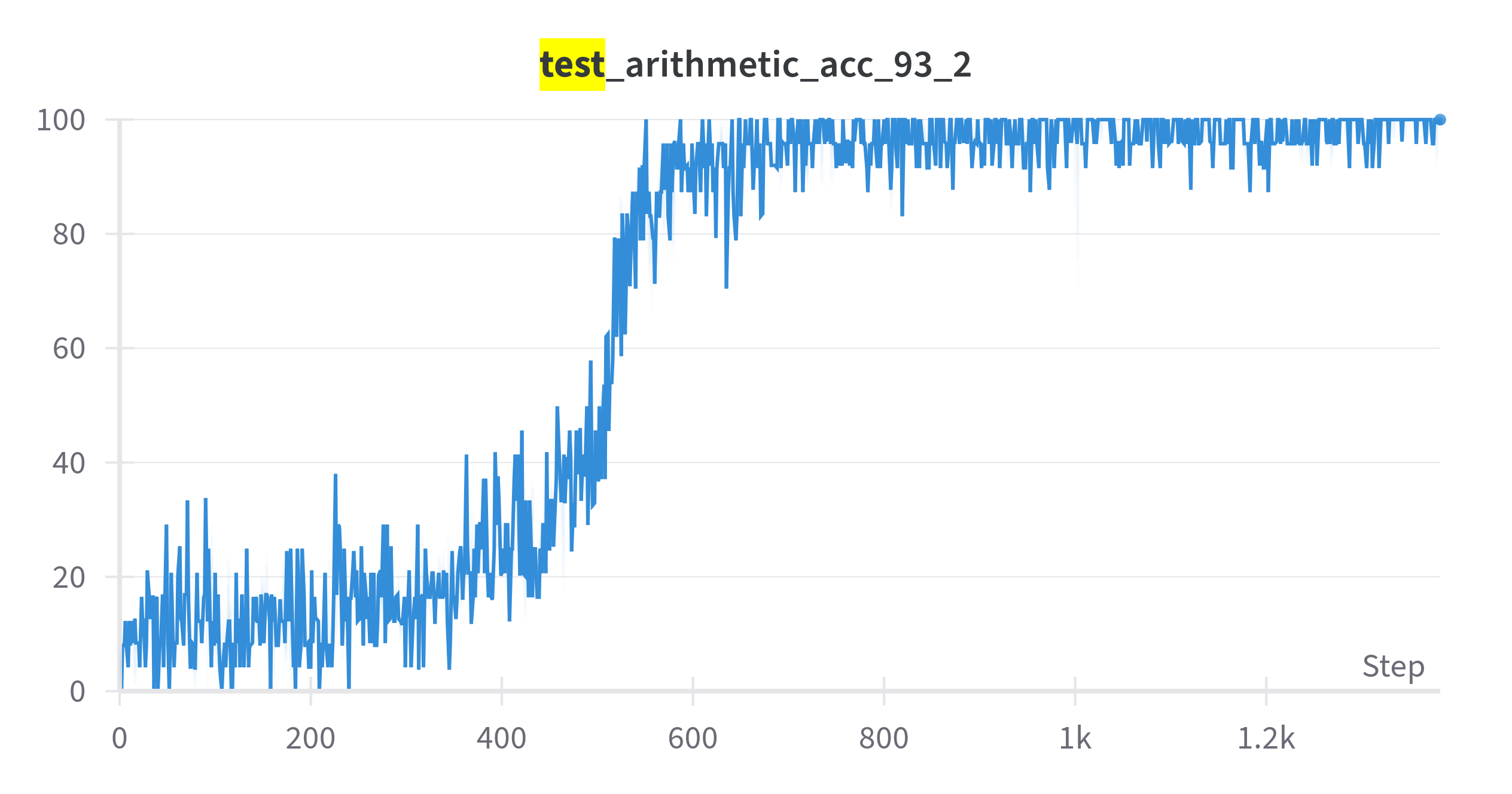}
    \end{minipage} \hfill
    \begin{minipage}[b]{0.30\linewidth}
        \includegraphics[width=\linewidth]{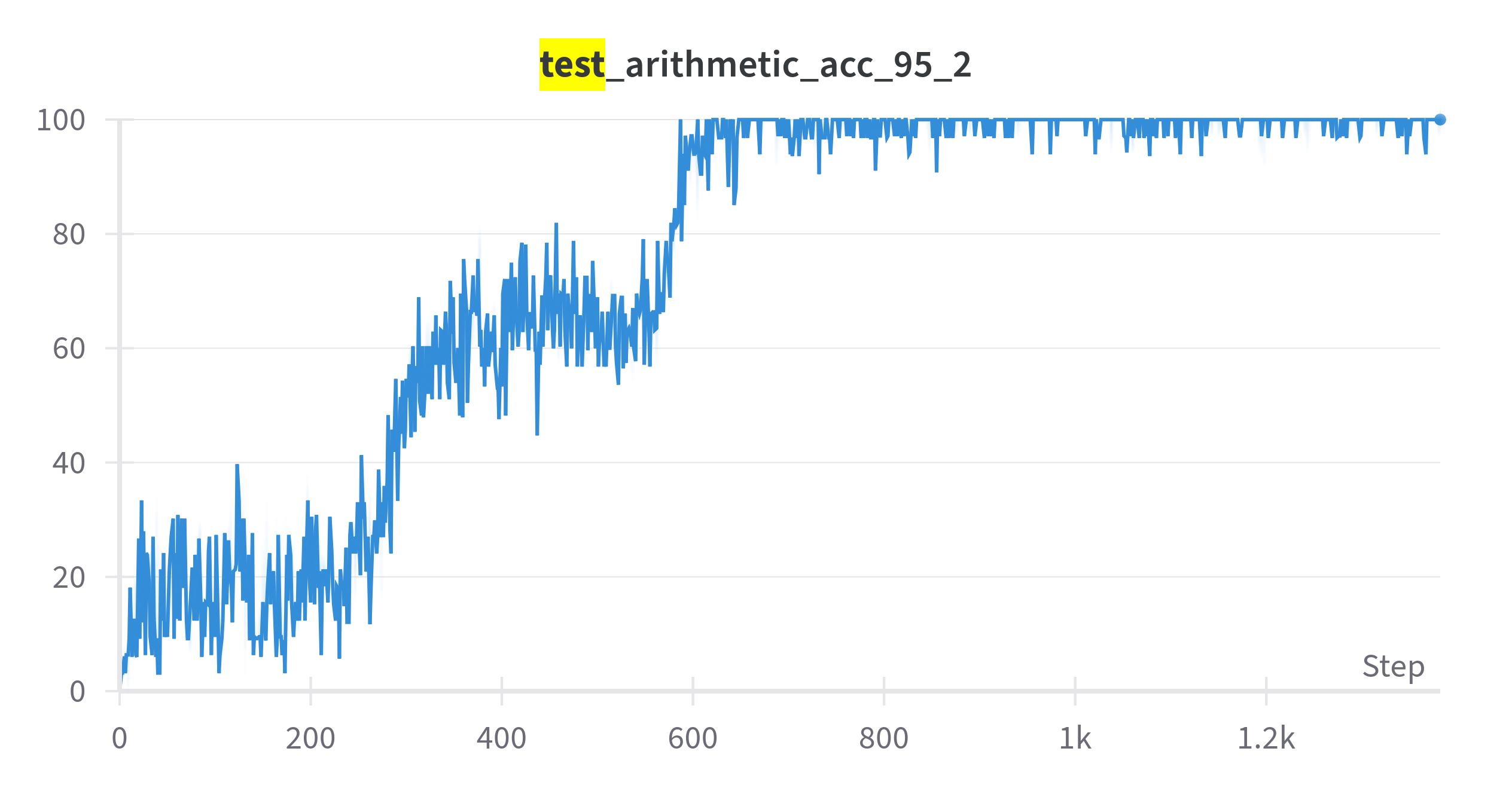}
    \end{minipage}

    \caption{Test plots from training runs for 18 different moduli.}
    \label{fig:valgrid_18}
\end{figure}

\end{document}